\documentclass[final,5p,times,twocolumn]{elsarticle}

\usepackage{amssymb}
\usepackage{amsmath}
\usepackage{bm}
\usepackage{color}
\def\red#1{\textcolor{red}{#1}}
\def\blue#1{\textcolor{blue}{#1}}

\newcommand{\argmax}{\mathop{\rm arg~max}\limits}

\usepackage[ruled, linesnumbered]{algorithm2e}

\usepackage{url}

\newcommand\blfootnote[1]{%
  \begingroup
  \renewcommand\thefootnote{}\footnote{#1}%
  \addtocounter{footnote}{-1}%
  \endgroup
}

\journal{Nuclear Physics B}

\begin{document}

\begin{frontmatter}

%% Title, authors and addresses

%% use the tnoteref command within \title for footnotes;
%% use the tnotetext command for theassociated footnote;
%% use the fnref command within \author or \affiliation for footnotes;
%% use the fntext command for theassociated footnote;
%% use the corref command within \author for corresponding author footnotes;
%% use the cortext command for theassociated footnote;
%% use the ead command for the email address,
%% and the form \ead[url] for the home page:
%% \title{Title\tnoteref{label1}}
%% \tnotetext[label1]{}
%% \author{Name\corref{cor1}\fnref{label2}}
%% \ead{email address}
%% \ead[url]{home page}
%% \fntext[label2]{}
%% \cortext[cor1]{}
%% \affiliation{organization={},
%%             addressline={},
%%             city={},
%%             postcode={},
%%             state={},
%%             country={}}
%% \fntext[label3]{}

\title{Human-in-the-loop Adaptation\\ in Group Activity Feature Learning for Team Sports Video Retrieval}

%% use optional labels to link authors explicitly to addresses:
% \author[label1,label2]{}
\author[label1]{Chihiro Nakatani}
\author[label2]{Hiroaki Kawashima}
\author[label1]{Norimichi Ukita}

\affiliation[label1]{organization={Toyota Technological Institute},
            addressline={2-12-1 Hisakata, Tempaku-ku},
            city={Nagoya},
            postcode={468-8511},
            state={Aichi},
            country={Japan}}

\affiliation[label2]{organization={University of Hyogo},
            addressline={8-2-1 Gakuennishi-machi, Nishi-ku},
            city={Kobe},
            postcode={651-2197},
            state={Hyogo},
            country={Japan}}

% \author{} %% Author name

%% Author affiliation
% \affiliation{organization={},%Department and Organization
%             addressline={}, 
%             city={},
%             postcode={}, 
%             state={},
%             country={}}

%% Abstract
\begin{abstract}
This paper proposes human-in-the-loop adaptation for Group Activity Feature Learning (GAFL) without group activity annotations. This human-in-the-loop adaptation is employed in a group-activity video retrieval framework to improve its retrieval performance. Our method initially pre-trains the GAF space based on the similarity of group activities in a self-supervised manner, unlike prior work that classifies videos into pre-defined group activity classes in a supervised learning manner. Our interactive fine-tuning process updates the GAF space to allow a user to better retrieve videos similar to query videos given by the user. In this fine-tuning, our proposed data-efficient video selection process provides several videos, which are selected from a video database, to the user in order to manually label these videos as positive or negative. These labeled videos are used to update (i.e., fine-tune) the GAF space, so that the positive and negative videos move closer to and farther away from the query videos through contrastive learning. Our comprehensive experimental results on two team sports datasets validate that our method significantly improves the retrieval performance. Ablation studies also demonstrate that several components in our human-in-the-loop adaptation contribute to the improvement of the retrieval performance. Code: \url{https://github.com/chihina/GAFL-FINE-CVIU}.

\end{abstract}

%%Graphical abstract
% \begin{graphicalabstract}
%\includegraphics{grabs}
% \end{graphicalabstract}

%%Research highlights
% \begin{highlights}
% \item Research highlight 1
% \item Research highlight 2
% \end{highlights}

%% Keywords
\begin{keyword}
%% keywords here, in the form: keyword \sep keyword
Group activity feature learning \sep Fine-tuning \sep Human-in-the-loop \sep Team sports video retrieval \sep

%% PACS codes here, in the form: \PACS code \sep code

%% MSC codes here, in the form: \MSC code \sep code
%% or \MSC[2008] code \sep code (2000 is the default)

\end{keyword}

\end{frontmatter}

%% Add \usepackage{lineno} before \begin{document} and uncomment 
%% following line to enable line numbers
%% \linenumbers

%% main text
%%

%%%%%%%%%%%%%%%%%%%%%%%%%%%%%%%%%%%%%%%%%%%%%%%%%%

\blfootnote{This is an accepted manuscript of an article published by Elsevier in Computer Vision and Image Understanding on 2026/01, available online: \url{https://doi.org/10.1016/j.cviu.2025.104577}}

\section{Introduction}
\label{sec:introduction}

\begin{figure}[t]
\begin{center}
\includegraphics[width=\columnwidth]{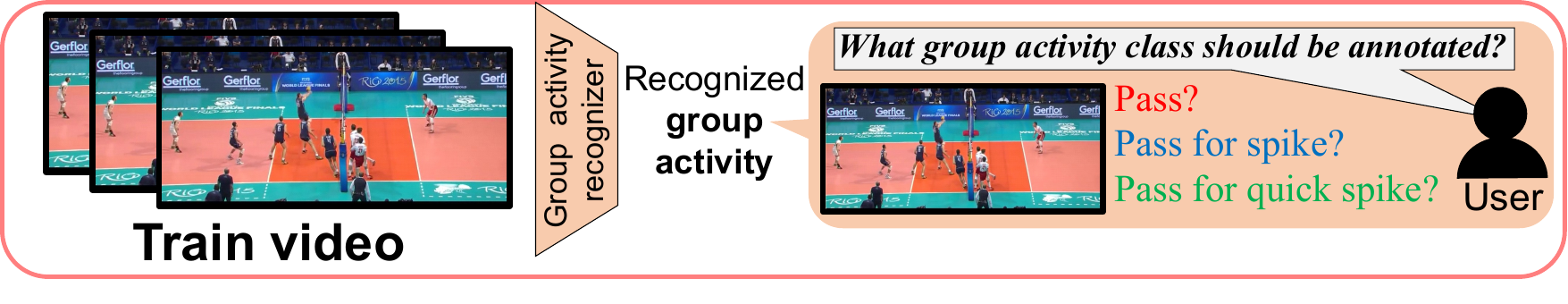}
(a) Supervised group activity recognition
\includegraphics[width=\columnwidth]{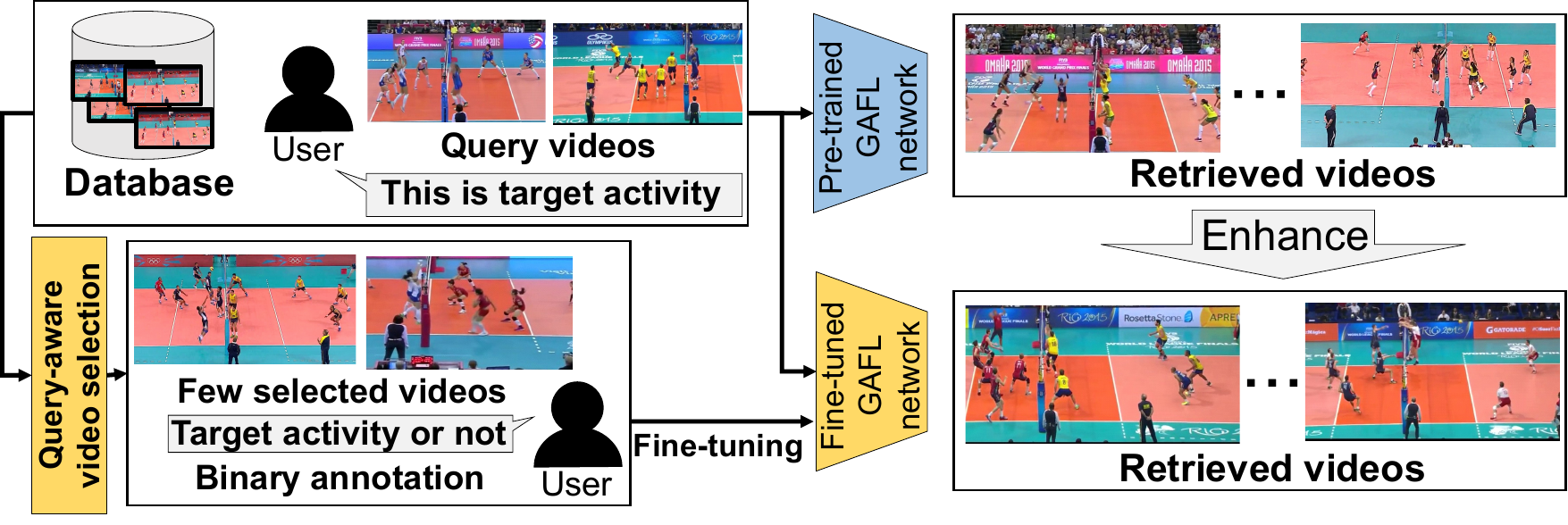}
(b) Our adaptive fine-tuning of group activity feature learning
\end{center}
\caption{
% Difference between previous method and our adaptive fine-tuning of group activity feature learning.
Difference between previous method and our method.
(a) Supervised GAR employs group activity annotations that are difficult due to various similar group activities.
(b) Our proposed method adaptively fine-tunes the pre-trained GAF space for retrieving target group activity represented by query videos given by users. 
Our fine-tuning only requires binary annotations to users on a few selected videos, eliminating group activity annotations.
}
\label{fig:top}
\end{figure}

% Group activities, characterized by multiple people engaging in interactions within the same group, are important cues for team sports analysis.
Group activities are characterized by multiple people engaging in interactions within the same group. 
The group activities are important cues for team sports analysis.
Group Activity Recognition (GAR) is a task to classify a frame or a video into predefined group activity classes~\cite{DBLP:conf/eccv/IbrahimM18,DBLP:conf/cvpr/WuWWGW19,DBLP:conf/cvpr/AzarANA19,DBLP:conf/eccv/PramonoCF20,DBLP:conf/eccv/EhsanpourASSRR20,DBLP:conf/cvpr/GavrilyukSJS20,DBLP:conf/iccv/Yuan0W21,DBLP:conf/iccv/LiCLYLHY21,DBLP:conf/cvpr/0002Z0Y0CQ22,DBLP:conf/eccv/TamuraVV22,DBLP:conf/eccv/ZhouKSGLZLKG22,DBLP:conf/cvpr/KimLCK22,DBLP:conf/iccvw/ChoiSS09,DBLP:conf/eccv/YanXTS020,DBLP:conf/mva/NakataniSU21,DBLP:conf/cvpr/XieGWC23,DBLP:journals/pami/YanXTST23,DBLP:conf/eccv/LiCLS24,DBLP:conf/eccv/KimSCK24,DBLP:conf/cvpr/ZhangLXZW24,DBLP:conf/eccv/NugrohoWLPWKK24}. 
These GAR methods are trained in a supervised manner, which requires a large number of ground-truth group activity annotations, as shown in Fig.~\ref{fig:top} (a).

% However, compared to the actions of a single person, annotating group activities is more labor-intensive and less reliable due to a large number of group people (e.g., 12 and 10 in volleyball and basketball, respectively). 
Annotating group activities is more challenging than a single person's actions. In a single person's action recognition, AI-driven wearable systems~\cite{Nyangaresi_Shanshool_2024} allow for constant tracking of the target person's activity. However, group activities with a large number of group people (e.g., 12 and 10 in volleyball and basketball, respectively) make the annotations more labor-intensive and less reliable.
Furthermore, to apply GAR to complex real-world applications (e.g., team sports analysis), it is not easy to appropriately pre-define all group activities as follows: (1) complex interactions among people make it difficult to define all group activities necessarily and sufficiently and (2) the necessary and sufficient classes of group activities may differ depending on various factors (e.g., users, scenarios, times, and applications). 
For example, 200 or more plays are defined every year in American Football games~\cite{bib:AFP2016,bib:AFP2021,bib:NFLP2018}.

To avoid these difficulties in annotating group activities, Group Activity Feature Learning (GAFL)~\cite{DBLP:conf/cvpr/NakataniKU24} trains Group Activity Features (GAFs) in a self-supervised manner without group activity annotations.
% , as shown in Fig.~\ref{fig:top} (b). 
%
%The trained Group Activity Feature (GAF) can be directly applied to similar video retrieval with querie videos.
%While the GAF trained by GAFL suggests the potential of self-supervised learning for complex group activities, the GAF is still insufficient for downstream tasks (e.g., retrieval using the GAF).
While GAFL provides no group activity classes unlike supervised GAR, the trained GAF is directly applicable to video retrieval from a video database.
%with query videos given by a user.
This retrieval is useful for strategy and tactics analysis in team sports.
% 
% For example, retrieving videos where a specific spike pattern leading to points is observed can help tactics analysis. 
% 
For example, retrieving videos that show a specific spike pattern leading to points can help tactics analysis. 

\begin{figure}[t]
\begin{center}
\includegraphics[width=\columnwidth]{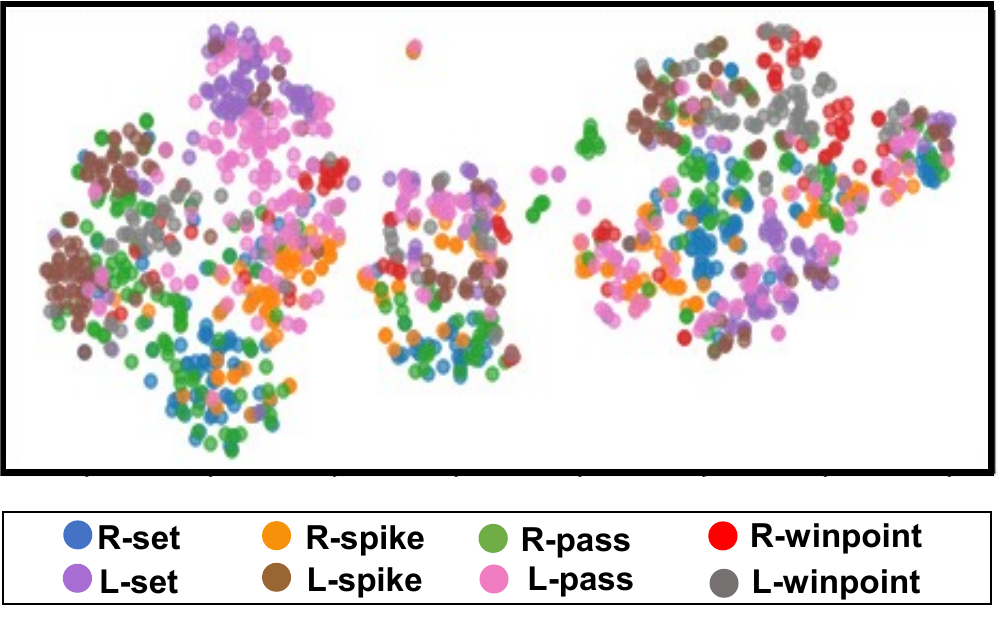}
\end{center}
\caption{
Group activity feature space learned by~\cite{DBLP:conf/cvpr/NakataniKU24} in a self-supervised manner on the Volleyball dataset~\cite{DBLP:conf/cvpr/IbrahimMDVM16}. 
% Each point denotes the group activity feature extracted from a video. 
% The group activity feature is transformed into 2-dimensional features by t-SNE~\cite{van2008visualizing}. 
The group activity feature extracted from each video is transformed into 2-dimensional features by t-SNE~\cite{van2008visualizing}. 
The color of each point corresponds to its group activity label annoated in the Volleyball dataset. 
While the manually annotated group activity labels are not used for the training, the group activity labels are only used for the visualization in this figure. The visualization reveals that group activity features are not learned enough in~\cite{DBLP:conf/cvpr/NakataniKU24} to discriminate between the different group activities.
}
\label{fig:gaf_space_prev}
\end{figure}

%While GAFL suggests the potential of self-supervised learning for discriminating complex group activities,
However, GAF trained only with a self-supervised manner is still insufficient.
For example, the GAF trained by GAFL on the Volleyball dataset~\cite{DBLP:conf/cvpr/IbrahimMDVM16} is visualized in Fig.~\ref{fig:gaf_space_prev} with t-SNE~\cite{van2008visualizing}. 
In this example, the color of each point corresponds to its manually annotated group activity label. This visualization reveals that the learned GAF is not enough to discriminate between group activities, as many points with different colors are mixed.

% Since self-supervised learning is in general insufficient, many self-supervised learning methods are expected to provide pre-trained models that are fine-tuned with a manually annotated dataset for a specific downstream task. The GAF can also be used as such a pre-trained model fine-tuned in a supervised manner. Specifically, various methods are proposed for supervised GAR~\cite{DBLP:conf/eccv/IbrahimM18,DBLP:conf/cvpr/WuWWGW19,DBLP:conf/cvpr/AzarANA19,DBLP:conf/eccv/PramonoCF20,DBLP:conf/eccv/EhsanpourASSRR20,DBLP:conf/cvpr/GavrilyukSJS20,DBLP:conf/iccv/Yuan0W21,DBLP:conf/iccv/LiCLYLHY21,DBLP:conf/cvpr/0002Z0Y0CQ22,DBLP:conf/eccv/TamuraVV22,DBLP:conf/eccv/ZhouKSGLZLKG22,DBLP:conf/cvpr/KimLCK22,DBLP:conf/iccvw/ChoiSS09,DBLP:conf/eccv/YanXTS020,DBLP:conf/mva/NakataniSU21,DBLP:conf/cvpr/XieGWC23,DBLP:journals/pami/YanXTST23,DBLP:conf/eccv/LiCLS24,DBLP:conf/eccv/KimSCK24,DBLP:conf/cvpr/ZhangLXZW24,DBLP:conf/eccv/NugrohoWLPWKK24}.
% 
% However, it is not easy to appropriately define and annotate a variety of complex group activities due to the two issues mentioned above (i.e., (1) difficulty in defining all group activities and (2) difficulty in adapting to various definitions of group activities).

%Rather than supervised GAR, as another prospective downstream task that can avoid these two issues, this paper focuses on a retrieval task. Our method fine-tunes the pre-trained GAF space for the retrieval task by human-in-the-loop adaptation based on query videos given by users (Fig.~\ref{fig:top} (c)).
% 
To improve GAFL, our method incorporates human-in-the-loop adaptation with GAFL.
% 
% To improve GAFL, this paper proposes a novel human-in-the-loop adaptation for retrieval, as with~\cite{Awad_Tulaib_Saleh_2024}, in which domain-specific adaptation yields novel contributions within existing biometrics frameworks.
% 
% Our method fine-tunes the GAF space pre-trained in a self-supervised manner to better retrieve videos similar to query videos given by a user in a human-in-the-loop manner (Fig.~\ref{fig:top} (b)).
% 
Our method fine-tunes the GAF space, which was pre-trained in a self-supervised manner. 
This fine-tuning aims to better retrieve videos similar to user-provided query videos in a human-in-the-loop manner (Fig.~\ref{fig:top} (b)).
% 
% The user simply provides a few query videos with no annotations of group activity labels, instead of defining group activity classes linguistically and semantically, which are complex as mentioned above.
The user simply provides a few query videos with no annotations of group activity labels.
This approach avoids the need to define group activity classes linguistically and semantically, which are inherently complex as mentioned above.
%%%Note that the user does not have to clearly define the target group activity class linguistically and semantically. All the users have to do is intuitively choose query videos similar to what they want to retrieve.

Using the query videos, our proposed video selection process chooses several videos from a video database.
Unlike general uncertainty-based selection in active learning~\cite{DBLP:conf/cvpr/LiG13,DBLP:conf/wacv/TaketsuguU24,DBLP:conf/cvpr/YangHC24}, this paper proposes query-aware video selection specifically designed for Group Activity Retrieval (GARet).
The user annotates whether these selected videos are what to expect as group activity videos or not (i.e., positive or negative).
That is, different from supervised GAR methods where annotators need to classify a large number of training videos into many complex group activity classes, our method simply and directly requires the user to judge whether or not each selected video matches their target group activity videos.
Finally, the GAF space is fine-tuned by the annotated selected videos to improve the GAF discriminability on videos similar to the users' query videos.

% For efficiently selecting videos for fine-tuning, we propose query-aware video selection.
% % 
% Our proposed video selection consists of two steps.
% % 
% First, several videos (e.g., 30) are selected from the training videos based on the similarity between each query video.
% % 
% These videos are important for fine-tuning because they include hard-negative samples (i.e., videos close to the query video, but with a different group activity from the query video).
% % 
% While annotating these selected videos can enhance retrieval performance with query videos, the selection relies on the GAF similarity from each query video, which can easily lead to selecting similar videos.
% % 
% To address the problem, we further apply a clustering-based selection, in which representative videos are identified from the selected videos.

% This query-based fine-tuning framework improves the performance of the retrieval task using the GAF model while avoiding the two issues mentioned above because (1) no pre-defined group activity classes are required and (2) the GAF model can be easily optimized for each user's task at test time.
%
% The retrieval task is important in sports analysis~\cite{DBLP:journals/tcsv/Shih18,DBLP:journals/spm/KokaramRDTBGS06}. 
% 
% For example, collecting videos where a specific spike pattern leading to points is observed can help tactics analysis. 
% 
Since the user demands (e.g., what kind of spike patterns are required) vary on a case-by-case basis, retrieving a particular target group activity is essential.
In terms of the user experience and usability in sports analysis applications, the amount of video annotations should be minimal. 
Thus, our human-in-the-loop adaptation requests the user to label only a limited number of selected videos (e.g., five) once as positive or negative.

For example, our human-in-the-loop adaptation supports a use case as follows: (i) an analyst provides a few query videos representing the target group activity, such as sample videos in which the team gets a point; (ii) the analyst annotates videos selected from a database of past plays; and (iii) the analyst receives retrieved videos from the fine-tuned GAF model to analyze tactics leading to points based on many videos.

Our novel contributions are summarized as follows:
\begin{itemize}
\item {\bf Human-in-the-loop adaptation in GAFL:}
%Unlike supervised GAR with group activity annotations, 
%which are not suitable for analyzing team sports,
This paper proposes a human-in-the-loop adaptation scheme for GARet.
In this scheme, a pre-trained GAF space~\cite{DBLP:conf/cvpr/NakataniKU24} is updated to improve the discriminativeness of a target group activity without group activity annotations.

\item {\bf Query-aware video selection for fine-tuning:}
For data-efficient fine-tuning, this paper proposes video selection using query videos. 
% 
%The key idea of this selection is that videos close to query videos are important because they help learn the subtle distinctions between target and non-target group activities.

\noindent{\bf Query similarity:}
    To select videos similar to query videos in terms of group activities, we select videos that are close to one of the query videos in the GAF space.
    Unlike active learning~\cite{DBLP:conf/cvpr/LiG13,DBLP:conf/wacv/TaketsuguU24,DBLP:conf/cvpr/YangHC24} using only the uncertainty of each unlabeled sample, we focus on the similarity between unlabeled videos and query videos to learn subtle distinctions between target and other group activities.

\noindent{\bf Query local dissimilarity:}
%    While the above query-similarity scoring is useful,
    %for data-efficient fine-tuning, it may select videos that are too similar and therefore meaningless for fine-tuning. To find videos that are similar but locally distinct from query videos, we propose to randomly perturb the query videos. This perturbation allows us to select important unlabeled videos that exhibit a large change of GAF similarity during this perturbation, indicating their local dissimilarity to the query videos.
    Sample video selection only with the above query similarity may select too similar videos, leading to inefficient fine-tuning. To avoid this problem, in evaluating the similarity between the query and sample videos, several people are masked to assess local dissimilarity (i.e., similarity of some people).
    %in addition to global similarity (i.e., similarity of all people).
    Sample video selection using this local dissimilarity in conjunction with the aforementioned query similarity (i.e., similarity of all people) allows us to select globally similar but locally dissimilar videos.

\item {\bf Comprehensive evaluation:}
We conduct comprehensive experiments on team sports datasets (i.e., volleyball~\cite{DBLP:conf/cvpr/IbrahimMDVM16} and basketball~\cite{DBLP:conf/eccv/YanXTS020}) to validate the applicability of our method in the team sports domain. 
% the domain of various team sports. 
%
Ablation studies also demonstrate that our method efficiently updates the pre-trained GAF space for the target group activity.
\end{itemize}

%%%%%%%%%%%%%%%%%%%%%%%%%%%%%%%%%%%%%%%%%%%%%%%%%%%%%%%%%%%%%%%%%%%%%%

\begin{figure*}
    \centering
    \includegraphics[width=\textwidth]{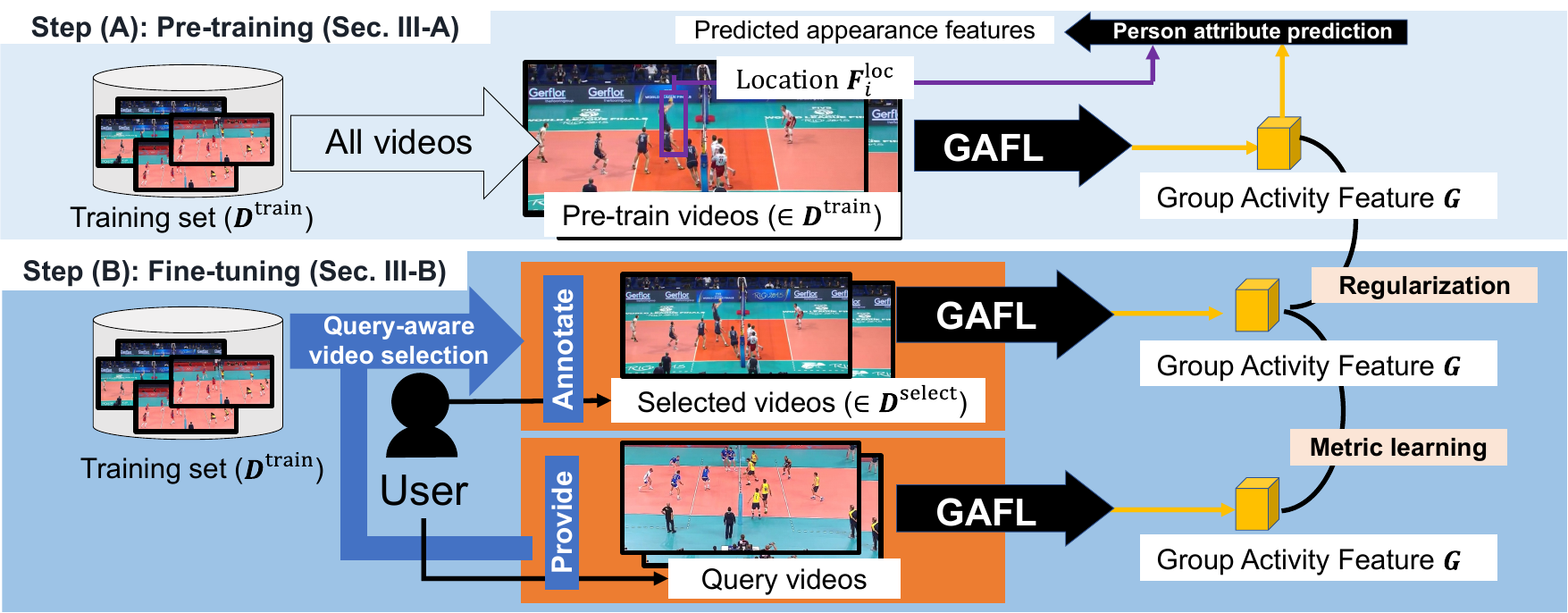}
    \caption{Overview of our proposal. (a) group activity feature learning network is pre-trained by~\cite{DBLP:conf/cvpr/NakataniKU24} with the training dataset. (b) The pre-trained group activity feature learning network is fine-tuned for target group activity presented by query videos given by users.}
    \label{fig:overview_gafl_fine}
\end{figure*}

\section{Related Work}
\label{sec:related_work}

\subsection{Group Activity Recognition (GAR)}
\label{subsec:gar}

%As introduced in Sec.~\ref{sec:introduction}, a frame of video is classified into one of the pre-defined group activity classes in GAR methods. The GAR network is trained in a supervised manner with manually annotated group activities.

In~\cite{DBLP:conf/iccv/Yuan0W21,DBLP:conf/cvpr/KimLCK22,DBLP:journals/pami/YanXTST23,DBLP:conf/eccv/NugrohoWLPWKK24,DBLP:conf/eccv/Li24}, manual group activity annotations are used for training GAR models.
In~\cite{DBLP:conf/cvpr/KimLCK22}, a group activity is recognized from a whole image. 
In~\cite{DBLP:conf/eccv/Li24}, human poses and object keypoints are fed into a group activity recognizor in which interaction between multi-person and object is built as a graph.
The set of person features is employed for a Graph Neural Network (GNN) by Kim~\textit{et al}~\cite{DBLP:conf/iccv/Yuan0W21} and Yan~\textit{et al}~\cite{DBLP:journals/pami/YanXTST23} as 
with~\cite{DBLP:conf/cvpr/WuWWGW19,DBLP:conf/eccv/EhsanpourASSRR20}.
%%%In~\cite{DBLP:conf/cvpr/WuWWGW19,DBLP:conf/eccv/EhsanpourASSRR20}, GNN in which each person and their interaction is regarded as node and edge, respectively, models the complex interactions of multiple people. 
In~\cite{DBLP:conf/cvpr/GavrilyukSJS20,DBLP:conf/iccv/LiCLYLHY21,DBLP:conf/cvpr/0002Z0Y0CQ22,DBLP:conf/eccv/TamuraVV22,DBLP:conf/eccv/ZhouKSGLZLKG22,DBLP:conf/cvpr/ZhangLXZW24}, Transformer~\cite{DBLP:conf/nips/VaswaniSPUJGKP17} improves the interaction modeling using a self-attention mechanism.
In~\cite{DBLP:conf/eccv/NugrohoWLPWKK24}, an optical flow map is used to guide attention maps in the transformer encoder to focus on foreground objects. 
 
% Following multi-task learning, the GAR network can be jointly trained with the person action recognition in~\cite{DBLP:conf/eccv/IbrahimM18,DBLP:conf/cvpr/WuWWGW19,DBLP:conf/cvpr/AzarANA19,DBLP:conf/eccv/PramonoCF20,DBLP:conf/eccv/EhsanpourASSRR20,DBLP:conf/cvpr/GavrilyukSJS20,DBLP:conf/iccv/LiCLYLHY21,DBLP:conf/cvpr/0002Z0Y0CQ22,DBLP:conf/eccv/TamuraVV22,DBLP:conf/eccv/ZhouKSGLZLKG22,DBLP:conf/iccvw/ChoiSS09,DBLP:conf/mva/NakataniSU21,DBLP:conf/cvpr/ZhangLXZW24} to augment the recognition performance of group activities.
% 
% In~\cite{DBLP:conf/eccv/Kim24}, the GAR network is jointly trained with people grouping. The network can recognize multiple group and their group activity classes at the same time.

Different from these GAR methods that are trained with predefined group activity annotations, our method fine-tunes a pre-trained GAF space based on any target group activity represented by query videos given by users.

\subsection{Group Activity Feature Learning (GAFL)}
\label{subsec:rel_ga_rep}

While general representation learning methods extract low-level features for various downstream tasks such as general image/video classification and prediction, several representation learning methods~\cite{DBLP:conf/eccv/IbrahimM18,DBLP:conf/cvpr/NakataniKU24} focus on high-level group activity features extracted from multiple people for group-related tasks (e.g., group scene retrieval and clustering). 
In~\cite{DBLP:conf/eccv/IbrahimM18}, scene features of multiple people are learned in a self-supervised manner through GNN.
Nakatani~\textit{et al}.~\cite{DBLP:conf/cvpr/NakataniKU24} also aims to extract the features of multiple people in a self-supervised manner through person attribute prediction as a pretext task.

% In our method, any pre-trained GAF space provided by these self-supervised methods~\cite{DBLP:conf/cvpr/NakataniKU24} is fine-tuned for the retrieval task.
In our method, any pre-trained GAF space provided by these self-supervised methods~\cite{DBLP:conf/cvpr/NakataniKU24} is fine-tuned with human feedback for the retrieval task. The domain-specific adaptation of existing frameworks that we propose is novel, similar to how Awad~\textit{et al}~\cite{Awad_Tulaib_Saleh_2024} have yielded novel contributions within existing recognition frameworks through proposing human body-aware adaptation for biometrics.

\begin{figure*}[t]
\begin{center}
\includegraphics[width=\textwidth]{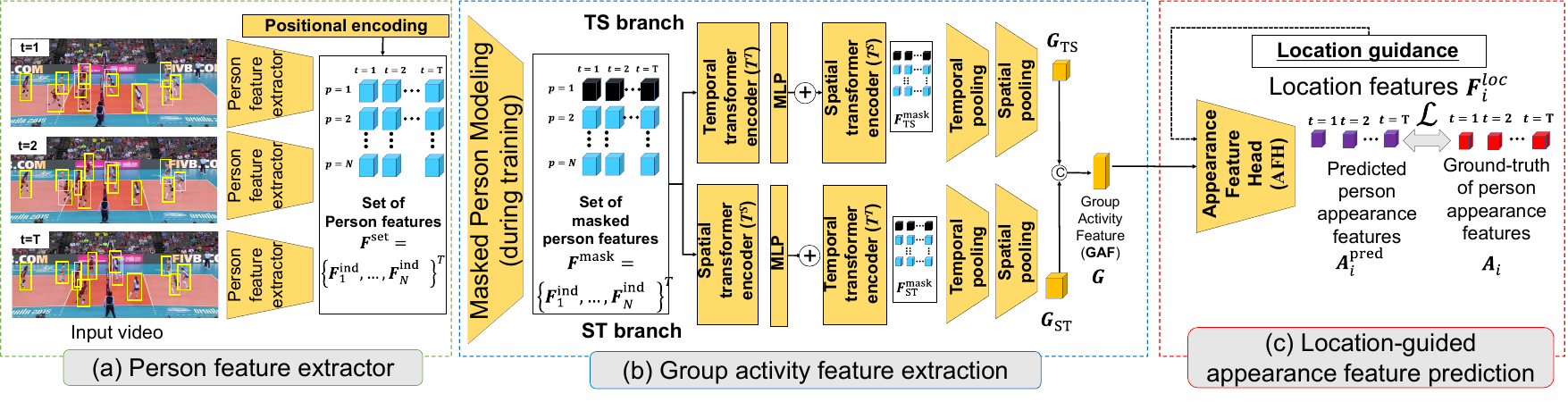}\\
\end{center}
\caption{Overview of our group activity feature learning network. (a) Person feature extractor. The person feature is composed of appearance and location features. (b) Group Activity Feature Learning (GAFL) network. The GAF is learned from extracted people features. (c) Location-guided appearance feature prediction network with the GAF. The appearance feature of each person is predicted from the location feature of the person and the GAF extracted in  (b). Through the appearance feature prediction, the GAF is learned in a self-supervised manner.
}
\label{fig:overview_network}
\end{figure*}

\subsection{Transfer Learning of Pre-trained Models}
\label{subsec:rel_trans_pre_train}

% As mentioned in Sec.~\ref{subsection:rel_ga_rep}, this paper aims to fine-tune the pre-trained GAF space for the retrieval task. 
% 
% For such adaptation of pre-trained models, many transfer learning methods have been investigated in recent decades.

\noindent{\bf Fine-tuning.}
%%%Fine-tuning is a well-known technique to adapt pre-trained models to downstream tasks. 
% 
One of the most important problems in fine-tuning is overfitting by catastrophic forgetting~\cite{DBLP:conf/iclr/0007WKWA21}. 
% 
%This problem motivates the community to investigate fine-tuning methods~\cite{DBLP:conf/icml/LiGD18,DBLP:conf/iclr/LiXWRLH19,DBLP:conf/nips/YouKL020,DBLP:conf/nips/LiuXX00JC022,DBLP:conf/iccv/HariharanG17} focusing on maintaining the pre-trained knowledge without overfitting. 
% 
To avoid this problem, in Davoine~\textit{et al}.~\cite{DBLP:conf/icml/LiGD18}, an L2 loss in which fine-tuned and pre-trained parameters are forced to be close is proposed.
Huan~\textit{et al}.~\cite{DBLP:conf/iclr/LiXWRLH19} propose fine-tuning in which
%the pre-trained model is updated while maintaining the similarity between feature vectors of the pre-trained and fine-tuned models.
the similarity between feature vectors of the pre-trained and fine-tuned models is maintained.
%  
%In Hariharan~\textit{et al}.~\cite{DBLP:conf/iccv/HariharanG17}, L1 or L2 losses are implemented to avoid extracting features useful only for small target datasets.
% 
\if 0
In Wang~\textit{et al}.~\cite{DBLP:conf/nips/YouKL020}, 
%the relationship between source and target datasets is learned. Then,
target labels are transferred to probabilistic labels in a source domain and used with the target labels for fine-tuning. 
\fi
% 
In Jin~\textit{et al}.~\cite{DBLP:conf/nips/LiuXX00JC022}, a subset of the source dataset is used for fine-tuning to avoid overfitting.

From the perspective of fine-tuning, our proposed method employs the regularization to avoid overfitting, as used in the previous methods~\cite{DBLP:conf/icml/LiGD18,DBLP:conf/iclr/LiXWRLH19}.

\noindent{\bf Sample selection.}
In active learning, samples that are informative for learning are selected to minimize annotation costs.
The criteria of sample selection can be divided into uncertainty-based selection and diversity-based selection.
The former considers the informativeness of each unlabeled sample for training, while the latter focuses on the dissimilarity between each unlabeled sample to avoid selecting similar samples.

For image classification, Li~\textit{et al}.~\cite{DBLP:conf/cvpr/LiG13} computes the entropy from the predicted image class probability and uses it as the uncertainty. In~\cite{DBLP:conf/wacv/TaketsuguU24}, for human pose estimation, uncertainty is computed from the pose heatmap.
%, which is the general representation for pose estimation.
Yang~\textit{et al}.~\cite{DBLP:conf/cvpr/YangHC24} propose an object detection-specific uncertainty that takes into account both classification and localization.
%difficulties.
%
Regarding diversity-based selection~\cite{DBLP:conf/iclr/SenerS18,DBLP:journals/corr/abs-1901-05954}, diverse samples are selected based on the mutual similarity between unlabelled samples. Core-set~\cite{DBLP:conf/iclr/SenerS18,DBLP:journals/corr/abs-1901-05954} selects a representative subset of an unlabeled dataset that covers the whole distribution of the dataset.
%While Zhdanov~\textit{et al}.~\cite{DBLP:journals/corr/abs-1901-05954} share the same idea with Core-set~\cite{DBLP:conf/iclr/SenerS18}, representative samples are selected by K-means clustering.

Similar to these methods, our method also focuses on reducing annotation costs.
Our method uses diversity-based selection, i.e., Core-set~\cite{DBLP:conf/iclr/SenerS18}, as one component.
On the other hand, uncertainty-based selection inevitably needs a model trained in a supervised manner, so it is not applicable to our method using a model pre-trained with no group activity annotation.
Instead of the uncertainty-based selection, we propose query-aware video selection specialized for GARet. 

\subsection{Team Sports Scene Retrieval}

Team sports scene retrieval is an important topic for team sports analysis, as proposed in~\cite{DBLP:journals/tmm/XuZZRLH08,DBLP:journals/mta/ChenC14,DBLP:conf/kdd/WangLCJ19, DBLP:journals/tkde/WangLC23,DBLP:conf/iui/ProbstKLRSSR18, DBLP:journals/tkdd/DiKSL18}.
Early approaches~\cite{DBLP:journals/tmm/XuZZRLH08,DBLP:journals/mta/ChenC14} describe each scene by the natural language representing the semantics of the scene (e.g., player A fails a free throw). 
While we can semantically retrieve a scene using the text as a query, it is challenging to perfectly represent the user's desired scene by text.
In trajectory-based retrieval~\cite{DBLP:conf/kdd/WangLCJ19, DBLP:journals/tkde/WangLC23,DBLP:conf/iui/ProbstKLRSSR18,DBLP:journals/tkdd/DiKSL18}, on the other hand, scene similarity is computed by comparing the trajectories of multiple players.
While real trajectories are used as a query in~\cite{DBLP:conf/kdd/WangLCJ19, DBLP:journals/tkde/WangLC23,DBLP:journals/tkdd/DiKSL18}, Probst~\textit{et al}~\cite{DBLP:conf/iui/ProbstKLRSSR18} propose to use a sketch-based query in which the users can represent their desired scene by drawing (e.g., drawing the trajectories of players).
Liu~\textit{et al}~\cite{DBLP:conf/IEEEcit/LiuH05} utilize visual features extracted from each video as a query for retrieval. 

% Difference 
% Similar to these methods, this paper also focuses on video retrieval in team sports. However, unlike these methods that utilize text and trajectory as queries, our method uses video itself as a query for retrieval.
Similar to these methods, this paper also focuses on retrieval in team sports. However, our method focuses on human-in-the-loop adaptation of the pre-trained GAF for query videos provided by the users.

%%%%%%%%%%%%%%%%%%%%%%%%%%%%%%%%%%%%%%%%%%%%%%%%%%%%%%%%%%%%%%%%%%%%%%

%%%%%%%%%%%%%%%%%%%%%%%%%%%%%%%%%%%%%%%%%%%%%%%%%%%%%%%%%%%%%%%%%%%%%%

\section{Proposed Method}
\label{sec:poposed_method}
  
Our method consists of the following two steps (Sec.~\ref{subsec:pretrain_gafl} and Sec.~\ref{subsec:gafl_finetune}, which are illustrated in Fig.~\ref{fig:overview_gafl_fine} (A) and Fig.~\ref{fig:overview_gafl_fine} (B), respectively).

In step (A), the GAF is pre-trained in a self-supervised manner~\cite{DBLP:conf/cvpr/NakataniKU24} with a training set $\bm{D}^{\mathrm{train}} = [D^{\mathrm{train}}_{1}, \cdots, D^{\mathrm{train}}_{N^{\mathrm{train}}}]$ where $N^{\mathrm{train}}$ denotes the number of training videos.
Although the novelty of our method does not lie in step (A), Sec.~\ref{subsec:pretrain_gafl} provides an overview of step (A), as our approach builds on the GAFL network pre-trained by step (A).

In step (B), which is proposed in this paper, the pre-trained GAFL network is updated to enhance the retrieval of the target group activity, represented by a few query videos (e.g., three) given by users only once.
Group activity videos are retrieved from $\bm{D}^{\mathrm{train}}$, which is used as a video dataset. 
The GAFL network is updated with a few videos (e.g., five) selected from $\bm{D}^{train}$ using our proposed query-aware video selection for fine-tuning.
Then, the users annotate the selected videos whether or not the target group activity is observed.
Using these annotated videos, metric learning is applied to the query videos to enhance the pre-trained GAF space with regularization for avoiding overfitting.

\if 0
As shown in Fig.~\ref{fig:overview_network}, the whole network for our GAF learning consists of three stages, (a), (b), and (c), which are introduced in detail in Sec.~\ref{subsec:pretrain_gafl}.
% 
In stage (a), the features of each person are extracted in each image independently. 
% 
In stage (b), the features of several people are masked (i.e., removed) during training for GAF enhancement in a self-supervised manner.
% 
The masked person features are fed into the transformer-based GAFL network. 
% 
In stage (c), to train the GAF space so that it represents the group-related activities of each person, the attribute of each person is predicted from the GAF with location guidance~\cite{DBLP:conf/cvpr/NakataniKU24}.
\fi
% 
% While the architecture of the GAFL network is the same in step (A) and step (B), the attribute prediction network is only used in the pre-training~\cite{DBLP:conf/cvpr/NakataniKU24}.

\subsection{Pre-training of GAFL network}
\label{subsec:pretrain_gafl}

% \subsubsection{Person Feature Extractor (stage (a))}
% \label{subsubsec:person_feature_extractor}
\noindent{\bf Person feature extractor (stage (a)).}
As shown in Fig.~\ref{fig:overview_network} (a), the set of person features $\bm{F}^{\mathrm{set}}_{k} \in\mathbb{R}^{T \times N \times C}$ are extracted from images of $k$-th video in $\bm{D}^{\mathrm{train}}$.
$\bm{F}^{\mathrm{set}}_{k}$ is composed of the $C$-dimensional features of $N$ people obtained from $T$ frames in the $k$-th video.
$\bm{F}^{\mathrm{ind}}_{k,i} \in\mathbb{R}^{C}$ is $i$-th person's feature vector indicated by a blue cuboid in Fig.~\ref{fig:overview_network} (a). 
$\bm{F}^{\mathrm{ind}}_{k,i}$ is obtained by the elementwise addition of appearance features (denoted by $\bm{F}^{app}_{k,i} \in\mathbb{R}^{C}$) and location features (denoted by $\bm{F}^{\mathrm{loc}}_{k,i} \in\mathbb{R}^{C}$).
The appearance feature $\bm{F}^{app}_{k,i}$ is obtained in three steps: (1) extracting a feature map from the whole image using VGG, (2) person feature map extraction from each person's bounding box by using RoIAlign, and (3) feature map embedding into a $C$-dimensional vector via a linear transformation.
$\bm{F}^{\mathrm{loc}}_{k,i}$ is obtained by a spatial positional encoding of the center point of each person bounding box, $(x, y)$, as with~\cite{DBLP:conf/cvpr/0002Z0Y0CQ22}.

% \subsubsection{Group Activity Feature Extraction (stage (b))}
% \label{subsubsec:group_feature_extraction}
\noindent{\bf Group activity feature extraction (stage (b)).}
To extract the features of people's interactions from 
% $\bm{F}^{\mathrm{set}}$,
$\bm{F}^{\mathrm{set}}_{k}$,
Masked Person Modeling (MPM) is applied to $\bm{F}^{\mathrm{set}}_{k}$ to obtain $\bm{F}^{\mathrm{mask}}_{k}$ during training.
Given $\bm{F}^{\mathrm{mask}}_{k}$, the GAFL network consists of two branches (i.e., temporal-to-spatial (TS) and spatial-to-temporal (ST) branches) similar to~\cite{DBLP:conf/cvpr/0002Z0Y0CQ22}. 
$\bm{F}^{\mathrm{mask}}_{k}$ is independently fed into the TS and ST branches to acquire 
$\bm{F}^{\mathrm{mask}}_{k,TS}$ and $\bm{F}^{\mathrm{mask}}_{k,ST}$, respectively.
Then, both temporal and spatial max pooling are applied to $\bm{F}^{\mathrm{mask}}_{k,TS}$ and $\bm{F}^{\mathrm{mask}}_{k,ST}$ to obtain $\bm{G}_{k,TS} \in\mathbb{R}^{C}$ and $\bm{G}_{k,ST}\in\mathbb{R}^{C}$, respectively.
Finally, $\bm{G}_{k,TS}$ and $\bm{G}_{k,ST}$ are concatenated to obtain the final GAF $\bm{G}_{k} \in\mathbb{R}^{2C}$.

% \subsubsection{Location-guided appearance feature prediction (stage~(c))}
% \label{subsubsec:loc_att_pred}
\noindent{\bf Location-guided appearance feature prediction (stage~(c)).}
During training, $\bm{G}_{k}$ is fed into the appearance feature head (denoted by $\mathrm{AFH}$), as shown in Fig.~\ref{fig:overview_network} (c). 
% 
% By backpropagating a loss used in this person attribute prediction not only inside $\mathrm{AFH}$ but also across the whole network shown in Fig.~\ref{fig:overview_network}, we can pre-train the GAFL network. 
% 
%%% Overview
% \paragraph{Overview.}
In $\mathrm{AFH}$, the appearance feature of each person (denoted by $\bm{A}_{k,i}^{\mathrm{pred}}\in\mathbb{R}^{T\times R}$) is predicted from $\bm{G}_{k}$ with their location features (i.e., $\bm{F}^{\mathrm{loc}}_{k,i}$ extracted in stage (a)) as guidance.
% Using $\bm{G}$, the appearance feature of each person is predicted by the $\mathrm{AFH}$ (Fig.~\ref{fig:overview_network} (c)). 
% 
% To predict the appearance feature of $i$-th person, their location features (i.e., $\bm{F}^{\mathrm{loc}}_{i}$) are used as guidance for attribute prediction from $\bm{G}$. 
% 
% The dimension of $R$ corresponds to the dimension of the appearance features (i.e., $C$). 
$R$ is the dimension of the person appearance features (i.e., $R = C$ in our method). 
% 
% While both the person action and person appearance features are used in~\cite{DBLP:conf/cvpr/NakataniKU24}, we only employ person appearance features because they require no manual annotations.
% 
The whole network is pre-trained in an end-to-end manner with a loss function as follows:
\begin{align}
    % \mathcal{L}_{paf} = \mathcal{L}_{MSE}(\bm{A}_{i}^{\mathrm{pred}}, \bm{A}_{i})
    \mathcal{L}_{paf} = \frac{1}{N^{\mathrm{train}}\cdot N}\sum_{k}^{N^{\mathrm{train}}}\sum_{i}^{N}  \mathcal{L}_{MSE}(\bm{A}_{k,i}^{\mathrm{pred}}, \bm{A}_{k,i})
\end{align}
% $\mathcal{L}_{paf}$ is the mean squared loss function where $\bm{A}_{i}$ denotes the extracted appearance features (i.e., $\bm{A}_{i} = \bm{F}^{app}$).
$\mathcal{L}_{paf}$ is the mean squared loss function where $\bm{A}_{k,i}$ denotes the appearance features extracted in stage (a) (i.e., $\bm{A}_{k,i} = \bm{F}_{k,i}^{app}$).

%%%%%%%%%%

\subsection{Human-in-the-loop adaptation of the pre-trained GAFL network}
\label{subsec:gafl_finetune}

\if 0
Section~\ref{subsec:gafl_finetune} describes how our proposed method updates the pre-trained GAFL network introduced in Sec.~\ref{subsec:pretrain_gafl} to better retrieve videos of the target group activity represented by query videos.
% 
In our method, query videos are chosen by users so that these videos represent what kind of group activities should be retrieved.
% 
While any number of query videos can be given, our human-in-the-loop adaptation can work with only a few videos (e.g., three).
% 
Note that these users do not have to clearly define the target group activity class linguistically and semantically.
% 
All the users have to do is intuitively choose query videos similar to what they want to retrieve.
%
% The goal of our method is to retrieve target videos from $\bm{D}^{\mathrm{train}}$ based on group activity-aware video similarity in the GAF space, as shown in Fig.~\ref{fig:overview_gafl_fine} (B).
% 
To update the pre-trained GAFL network, several videos (e.g., five) are selected from $\bm{D}^{\mathrm{train}}$, and the users are required to label these selected videos as positive or negative in a human-in-the-loop manner.
% 
% As annotating a large number of videos (e.g., 100) by the users is not easy for real-world applications, we propose a novel video selection method that selects videos efficiently for this adaptation from $\bm{D}^{\mathrm{train}}$.
\fi

For the user-guided adaptation of the GAFL network, $N^{\mathrm{select}}$ videos are selected and given to the users for annotating these videos.
In our method, $N^{\mathrm{select}} \times N^{E}$ videos are initially selected based on query-aware video selection as described in Sec.~\ref{subsubsec:select_query}, diversity-aware video selection narrows down these videos to $N^{\mathrm{select}}$ videos as described in Sec.~\ref{subsubsec:select_div}.
$N^{E}$, where $N^{E} \geq 1$, denotes a coefficient to determine how many extra videos are given to diversity-aware video selection in order to select $N^{\mathrm{select}}$ diverse videos for efficiently GAFL fine-tuning.

\begin{algorithm}[t]
\caption{Query-aware video selection (Sec.~\ref{subsubsec:select_query})}
\label{alg:query_selection}
\SetAlgoLined
\KwIn{Training set $\bm{D}^{\mathrm{train}}$ and Query set $\bm{D}^{\mathrm{query}}$}
\KwOut{$\bm{D}^{\mathrm{ex}}$}
\BlankLine
$\bm{D}_{ex} \leftarrow \emptyset $ \;
\For{$k=1$ \KwTo $N^{\mathrm{query}}$}{
    $H_{k} \leftarrow [] $ \;
    \For{$k'=1$ \KwTo $|\bm{D}^{\mathrm{train}}|$}{
        $\bm{S}_{k,k'} \leftarrow F_{\mathrm{cos}} (\bm{G}_{k},\bm{G}_{k'})$ \;
        $\bm{\bar{S}}_{k,k'} \leftarrow \frac{1}{P} \sum_{p}^{P} \bm{S}_{k,k'}^{p}$ \;

        $\bm{V}_{k,k'} \leftarrow \frac{1}{P} \sum_{p}^{P} (\bm{S}_{k,k'}^{p} - \bm{\bar{S}}_{k,k'})^{2}$ \;
        $\bm{I}_{k,k'} \leftarrow \bm{S}_{k,k'} + \lambda\bm{V}_{k,k'}$ \;
        Append $(I_{k,k'}, k')$ to $H_{k}$ \;
    }
    $H_{\mathrm{selected}} \leftarrow \text{TopK}(H_{k}, K=N^{\mathrm{select}}*N^{\mathrm{E}})$ \;
    \For{$(I_{k,k'_{\text{selected}}}, k'_{\text{selected}})$ \textbf{in} $H_{\mathrm{selected}}$}{
        $\bm{D}_{ex} \leftarrow \bm{D}_{ex} \cup \{ \bm{D}^{\mathrm{train}}_{k'_{\text{selected}}} \}$ \;
        $\bm{D}^{\mathrm{train}} \leftarrow \bm{D}^{\mathrm{train}} \setminus \{ \bm{D}^{\mathrm{train}}_{k'_{\text{selected}}}  \}$\;
    }
}
\BlankLine
\Return $\bm{D}_{ex}$ \;
\end{algorithm}

\begin{algorithm}[t]
\caption{Diversity-aware video selection (Sec.~\ref{subsubsec:select_div})}
\label{alg:div_selection}
\SetAlgoLined
\KwIn{$\bm{D}^{\mathrm{ex}}$}
\KwOut{Selected videos $\bm{D}^{\mathrm{select}}$}
\BlankLine
$\bm{D}^{\mathrm{select}} \leftarrow \{ \text{Randomly select one video from } \bm{D}^{\mathrm{ex}} \}$\;
$\bm{D}^{\mathrm{ex}} \leftarrow \bm{D}^{\mathrm{ex}} \setminus \{ D^{\mathrm{select}}_{1} \}$\;

\For{$r=1$ \KwTo $N^{\mathrm{select}}-1$}{
    $\bm{D}^{\mathrm{select}}_{r} \leftarrow \bm{D}^{\mathrm{select}}$\; 
    $u \leftarrow \argmax_{i\in\bm{D}^{\mathrm{ex}}} \left( \min_{j\in\bm{D}^{\mathrm{select}}_{r}}  F_{\mathrm{cos}}(\bm{G}_{i}, \bm{G}_{j}) \right)$ \;
    $\bm{D}^{\mathrm{select}} \leftarrow \bm{D}^{\mathrm{select}} \cup \{ \bm{D}^{\mathrm{ex}}_{u} \}$\;
    $\bm{D}^{\mathrm{ex}} \leftarrow \bm{D}^{\mathrm{ex}} \setminus \{ \bm{D}^{\mathrm{ex}}_{u} \}$\;
}
\BlankLine
\Return{$\bm{D}^{\mathrm{select}}$}\;
\end{algorithm}

\subsubsection{Query-aware video selection}
\label{subsubsec:select_query}

Algorithm~\ref{alg:query_selection} shows the procedure of our query-aware video selection.
% 
%Given query videos, 
Our query-aware video selection builds on the key idea that videos close to query videos in the GAF space are useful for fine-tuning.
This is because these videos contain subtle distinctions between target and non-target group activities.
Our query-aware video selection is defined by the following two criteria, i.e., query similarity and query local dissimilarity.
% 
\if 0
From the query videos, $N^{\mathrm{select}} \times N^{E}$ videos are selected from $\bm{D}^{train}$. 
% 
$N^{\mathrm{select}}$ denotes the number of the selected videos. 
% 
$N^{E}$ is the coefficient to control how many videos are selected in the query-aware video selection.
% 
Finally, these $N^{\mathrm{select}} \times N^{E}$ videos are filtered into $N^{\mathrm{select}}$ videos by diversity-aware video selection described in Sec.~\ref{subsubsec:select_div}.
% 
% From the selected $N^{\mathrm{select}} * N^{E}$ videos, $N^{\mathrm{select}}$ videos are finally selected using the following diversity-aware video selection, as described in Sec.~\ref{subsubsec:select_div}
\fi

\noindent{\bf Query similarity:}
The query similarity is designed to select videos that are close to the query videos from $\bm{D}^{\mathrm{train}}$.
The query videos $\bm{D}^{\mathrm{query}} = [D^{\mathrm{query}}_{1}, \cdots, D^{query}_{N^{\mathrm{query}}}]$, in which the target group activity is observed, are provided by the users. 
% 
% Given a set of $N^{\mathrm{query}}$ query videos, each video in $\bm{D}^{\mathrm{train}}$ is scored by the GAF similarity between each query video as follows:
Given a set of $N^{\mathrm{query}}$ query videos, each video in $\bm{D}^{\mathrm{train}}$ is scored by the GAF similarity with each query video as follows:
\begin{align}
    \bm{S}_{k,k'} = F_{\mathrm{cos}} (\bm{G}_{k},\bm{G}_{k'}),
    \label{eq:q_sim}
\end{align}
where $F_{\mathrm{cos}}(\bm{G}_{k},\bm{G}_{k'})$ denotes the cosine similarity between the GAF of $k$-th query video and the GAF of $k'$-th video in $\bm{D}^{\mathrm{train}}$, which are denoted by $\bm{G}_{k}$ and $\bm{G}_{k'}$, respectively.
$\bm{G}_{k}$ and $\bm{G}_{k'}$ are extracted by the pre-trained GAFL network (Sec.~\ref{subsec:pretrain_gafl}).

\noindent{\bf Query local dissimilarity:}
%The query similarity mentioned above is useful for selecting videos that are better suited for learning the difference between target and non-target group activities. However, several selected videos may be too similar to each other and not be informative for fine-tuning.
% Several videos selected by our query similarity mentioned above may be too similar to each other and not be informative for fine-tuning.
Several videos selected by our query similarity mentioned above may be too similar to each query video and not be informative for fine-tuning.
To address this, selecting videos that are neither too high nor too low $\bm{S}_{k,k'}$ may be a straightforward approach, which leads to video selection from a wider distribution of videos.
However, this approach still works only based on global similarity (i.e., the spatial configuration and visual features of all people).
%(e.g., selecting videos in each of which a people configuration is shifted only a little from those in the query videos). 
On the other hand, local dissimilarities represented by not all but several people are also essential to discriminate between target and non-target group activities.
Such local dissimilarities may be neglected in global similarity evaluation using $\bm{S}_{k,k'}$.
Therefore, we also propose query local dissimilarity, focusing on selecting videos locally dissimilar to the query videos.
% Therefore, we also propose query local dissimilarity, representing the video diversity from query videos.
% This method aims to select videos where only some people are similar to the query videos.

This query local dissimilarity is represented by the variation of GAFs extracted from partially masked people.
First of all, $N^{V}$ people are randomly masked in each query video and fed into the pre-trained GAFL network to obtain a Locally Masked GAF (LM-GAF).
Then, the GAF similarities between the LM-GAF and GAFs extracted from all videos in $\bm{D}^{\mathrm{train}}$ are computed.
A set of these GAF similarities is computed $P$ times with different random masking patterns, where $P$ is the number of patterns of masked people.
The variance of the $P$ GAF similarities computed from $k'$-th video in $\bm{D}^{\mathrm{train}}$ is regarded as an indicator of how much this video is locally dissimilar to the $k$-th query video, as follows:
\begin{align}
    \bm{V}_{k,k'} =  \frac{1}{P} \sum_{p}^{P} (\bm{S}_{k,k'}^{p} - \bm{\bar{S}}_{k,k'})^{2}\\
    \bm{\bar{S}}_{k,k'} = \frac{1}{P} \sum_{p}^{P} \bm{S}_{k,k'}^{p},
    \label{eq:q_var}
\end{align}
where $\bm{V}_{k,k'}$ denotes the variance of the $P$ GAF similarities between $k$-th query video and $k'$-th video in $\bm{D}^{\mathrm{train}}$.
$\bm{S}_{k,k'}^{p}$, where $p \in \{ 1, \cdots, P \}$, denotes the GAF similarity between the $p$-th LM-GAF extracted from $k$-th query video and the GAF extracted from $k'$-th video in $\bm{D}^{\mathrm{train}}$. 
% 
%$\bm{\bar{S}}_{k,k'}$ is the mean of the $P$ GAF similarities.

Why can $\bm{V}_{k,k'}$ be regarded as a local dissimilarity score? If all people in two videos (i.e., a query video and a sample video) are almost the same with no local dissimilarity, $\bm{V}_{k,k'}$ becomes larger only due to the variation of LM-GAFs of a query video.
On the other hand, if some people differ between the two videos, this difference makes $\bm{V}_{k,k'}$ much larger.
Therefore, $\bm{V}_{k,k'}$ is regarded as the query local dissimilarity in our method.

\begin{figure}[t]
\begin{center}
\includegraphics[width=\columnwidth]{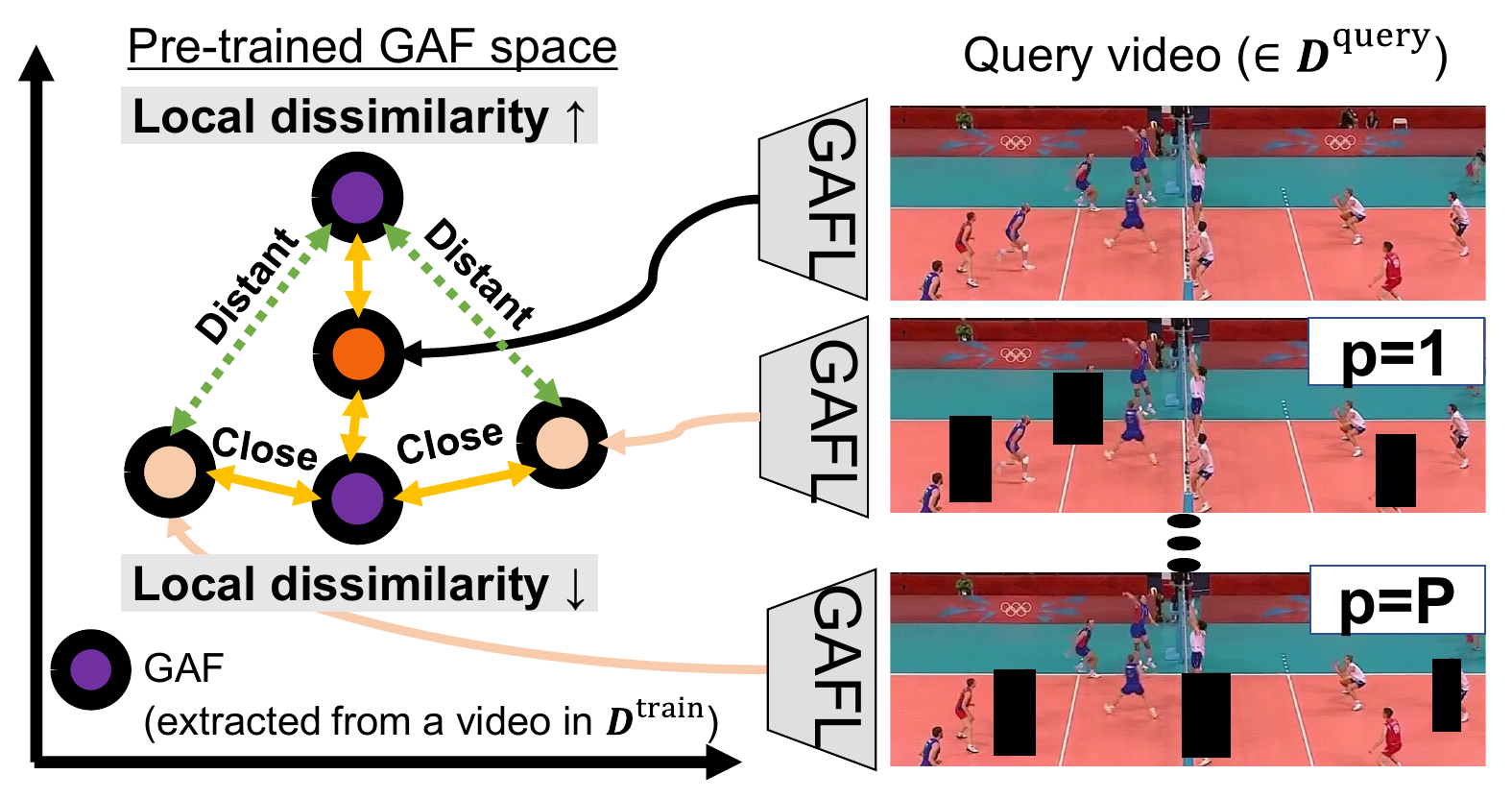}\\
\end{center}
\caption{Query local dissimilarity computation. The query GAF and Locally Masked query GAF are extracted by the pre-trained GAFL network. For each video in $\bm{D}^{\mathrm{train}}$, the variance of the GAF similarities between the GAFs extracted from the query videos during the masking is computed.
}
\label{fig:query_pert}
\end{figure}

\noindent{\bf Query-aware similarity:}
The query similarity (i.e., $\bm{S}_{k, k'}$) and query local dissimilarity (i.e., $\bm{V}_{k, k'}$) are ensembled to compute our informative score $\bm{I}_{k,k'}$. $\bm{I}_{k,k'}$ indicates how informative $k'$-th video in $\bm{D}^{\mathrm{train}}$ is for fine-tuning with respect to $k$-th query video as follows:
\begin{align}
    \bm{I}_{k,k'} = \bm{S}_{k,k'} + \lambda\bm{V}_{k,k'},
    \label{eq:q_info}
\end{align}
where $\lambda$ is the weight to balance $\bm{S}_{k,k'}$ and $\bm{V}_{k,k'}$.

For $k$-th query video, all videos in $\bm{D}^{\mathrm{train}}$ are sorted in ascending order of $\bm{I}_{k,k'}$.
Then, the top $N^{\mathrm{select}}\times N^{\mathrm{E}}$ videos are extracted for $k$-th query video.
In total, $N^{\mathrm{select}}\times N^{\mathrm{E}} \times N^{query}$ videos, denoted by $\bm{D}^{\mathrm{ex}} = [D^{\mathrm{ex}}_{1}, \cdots, D^{ex}_{N^{\mathrm{select}}\times N^{\mathrm{E}} \times N^{quey}}]$, are extracted from all videos in $\bm{D}^{\mathrm{train}}$.

\subsubsection{Diversity-aware video selection}
\label{subsubsec:select_div}

%In addition to query-aware video selection, which focuses on the informativeness of each video in $\bm{D}^{\mathrm{train}}$, we also employ diversity-aware video selection to prevent the selection of similar videos.
% 
%As mentioned in~\cite{DBLP:conf/iclr/SenerS18,DBLP:journals/corr/abs-1901-05954}, selecting mutually similar videos leads to inefficient fine-tuning in terms of the number of videos annotated by users.
% 
% While query local dissimilarity in our query-aware video selection improves the diversity of videos, our method also uses Core-Set~\cite{DBLP:conf/iclr/SenerS18} to obtain a representative subset that covers the distribution of the entire set of samples, for further diversity.
While our query-aware video selection improves the GAFL fine-tuning efficiency, our method also uses Core-Set~\cite{DBLP:conf/iclr/SenerS18} to obtain a representative subset that covers the distribution of the entire set of samples for diversity. 
Using Core-Set, $N^{\mathrm{select}}$ videos are selected from $\bm{D}^{\mathrm{ex}}$.
The procedure of the diversity-aware video selection using Core-Set is shown in Algorithm~\ref{alg:div_selection}.

According to Core-set, the first video (denoted by $D^{\mathrm{select}}_{1}$) is selected randomly from $\bm{D}^{\mathrm{ex}}$.
The second and later videos are selected iteratively.
The video selected in the $r$-th iteration is denoted by $D^{\mathrm{select}}_{r}$.
Given $\bm{D}^{\mathrm{select}}_{r}$, which is the set of selected videos by the $r$-th iteration, $D^{\mathrm{select}}_{r+1}$ is selected as follows:
\if 0
\begin{align}
    D^{\mathrm{select}}_{r+1} = \max_{i\in\bm{D}^{\mathrm{ex}}} \min_{j\in\bm{D}^{\mathrm{select}}_{r}}  F_{cos}(\bm{G}_{i}, \bm{G}_{j}).
    \label{eq:core_set}
\end{align}
\fi
\begin{align}
    D^{\mathrm{select}}_{r+1} &= D^{\mathrm{ex}}_{u}\\
    u &= \argmax_{i\in\bm{D}^{\mathrm{ex}}} \min_{j\in\bm{D}^{\mathrm{select}}_{r}}  F_{cos}(\bm{G}_{i}, \bm{G}_{j})
    \label{eq:core_set}
\end{align}
This video selection is looped $N^{\mathrm{select}}$ times to obtain $\bm{D}^{\mathrm{select}} = [D^{\mathrm{select}}_{1}, \cdots, D^{\mathrm{select}}_{\mathrm{N}_{select}}]$.

The users annotate these $N_{select}$ videos in $\bm{D}^{\mathrm{select}}$ as positive or negative for fine-tuning.
While the binary annotation can be done by one user, we can apply multi-user annotation to improve the annotation quality. For example, more than one user is asked to annotate these $N_{select}$ videos. Then, we can integrate the annotation results by multiple users in some way (e.g., majority voting) to improve the reliability of the annotation data.

\subsubsection{Loss functions}
\label{subsubsec:loss_func}

In our fine-tuning, the pre-trained GAF space is updated with contrastive learning~\cite{DBLP:conf/icml/ChenK0H20} using the positive and negative labels given to $\bm{D}^{\mathrm{select}}$, as mentioned at the beginning of Sec.~\ref{subsec:gafl_finetune}.
This contrastive learning is achieved with GAFs extracted from the query and selected videos.
MSE loss to maintain the pre-trained GAF space is also applied to avoid overfitting as proposed in~\cite{DBLP:conf/iclr/LiXWRLH19}.
The whole network is fine-tuned in an end-to-end manner with the following loss function, $\mathcal{L}$, as follows:
\begin{align}
    \mathcal{L} &= \mathcal{L}_{ctr} + \mathcal{L}_{reg}\\
    \mathcal{L}_{ctr} &= \frac{1}{N^{\mathrm{query}}}\sum_{k}^{N^{\mathrm{query}}}\max(0, d(\bm{G}_{k}, \bm{G}_{\mathrm{pos}}) - d(\bm{G}_{k}, \bm{G}_{\mathrm{neg}}) + \alpha)\\
    \mathcal{L}_{reg} &= \frac{1}{N^{\mathrm{select}}}\sum_{k'}^{N^{\mathrm{select}}}\mathcal{L}_{MSE}(\bm{G}_{k'},\bm{G}^{\mathrm{pretrain}}_{k'})
    \label{eq:finetune_loss}
\end{align} 
$\mathcal{L}_{ctr}$ is the triplet loss function where $\bm{G}_{\mathrm{pos}}$ and $\bm{G}_{\mathrm{neg}}$ denote the GAFs extracted from positive and negative selected videos, respectively. \textit{d} measures the Euclidean distance between two GAFs. $\alpha$ denotes the margin.
$\mathcal{L}_{reg}$ is a regularization term designed to preserve the pre-trained GAF space. $\bm{G}^{\mathrm{pretrain}}_{k'}$ denotes the GAF extracted from the $k'$-th selected video with the pre-trained GAFL network before fine-tuning.

%%%%%%%%%%%%%%%%%%%%%%%%%%%%%%%%%%%%%%%%%%%%%%%%%%%%%%%%%%%%%%%%%%%%%%

\begin{table}[t]
\caption{Distribution of group activity classes annotated in the Volleyball dataset.}
\label{table:det_ann_vol}
\centering
\begin{tabular}{l||r}\hline
Group activity class & Number of labels \\ \hline
Right-set              & 644              \\ \hline
Right-spike            & 623              \\ \hline
Right-pass             & 801              \\ \hline
Right-winpoint         & 295              \\ \hline
Left-winpoint          & 367              \\ \hline
Left-pass              & 826              \\ \hline
Left-spike             & 642              \\ \hline
Left-set               & 633              \\ \hline
\end{tabular}
\end{table}

\begin{table}[t]
\caption{Distribution of group activity classes annotated in the NBA dataset.}
\label{table:det_ann_nba}
\centering
\begin{tabular}{l||r}\hline
Group activity class & Number of labels \\ \hline
2p-succ.              & 961              \\ \hline
2p-fail-off.          & 541              \\ \hline
2p-fail-def.          & 1550             \\ \hline
2p-layup-succ.        & 994              \\ \hline
2p-layup-fail.-off    & 544              \\ \hline
2p-layup-fail.-def    & 859              \\ \hline
3p-succ.              & 911              \\ \hline
3p-fail.-off          & 602              \\ \hline
3p-fail.-def          & 2210             \\ \hline
\end{tabular}
\end{table}

\begin{table}[t]
\caption{Distribution of group activity classes annotated in the Collective Activity dataset.}
\label{table:det_ann_cad}
\centering
\begin{tabular}{l||r}\hline
Group activity class & Number of labels \\ \hline
Moving              & 1111              \\ \hline
Waiting          & 455              \\ \hline
Queuing          & 502             \\ \hline
Talking        & 450              \\ \hline
\end{tabular}
\end{table}

\begin{table*}[t]
    \centering
    \caption{
    Quantitative comparison of retrieval on the VolleyBall Dataset (Volleyball).
    The best result in each column is colored in \textcolor{red}{red}.
    In ``Ours'', the whole network is fine-tuned for each target group activity with five selected videos for fine-tuning. As query videos, three videos are sampled from the test videos for each fine-tuning. 
    The retrieval results are the weighted average of eight group activity classes, considering the number of videos in each group activity class.}
    \begin{tabular}{l||c|c|c|c|c|c|c|c} \hline
    & Precision@5 & Precision@10 & Precision@5 & Precision@10 & Hit@5 & Hit@10 & Hit@5 & Hit@10 \\
    & original & original & others & others & original & original & others & others \\ \hline
    B1-Compact128~\cite{DBLP:conf/eccv/IbrahimM18} & 0.318 & 0.299 & 0.315 & 0.293 & 0.737 & 0.870 & 0.761 & 0.900 \\ \hline
    B2-VGG19~\cite{DBLP:conf/eccv/IbrahimM18} & 0.319 & 0.291 & 0.322 & 0.297 & 0.791 & 0.905 & 0.777 & 0.907 \\ \hline
    HRN~\cite{DBLP:conf/eccv/IbrahimM18} & 0.280 & 0.254 & 0.276 & 0.262 & 0.728 & 0.861 & 0.728 & 0.881 \\ \hline
    GAFL~\cite{DBLP:conf/cvpr/NakataniKU24} & 0.557 & 0.533 & 0.557 & 0.535 & 0.886 & 0.962 & 0.881 & 0.936 \\ \hline
    Ours & \red{0.739} & \red{0.647} & \red{0.596} & \red{0.564} & \red{0.988} & \red{0.993} & \red{0.905} & \red{0.950} \\ \hline
    \end{tabular}
    \label{table:comp_vol}
\end{table*}

\section{Experiments}
\label{sec:experiments}

\subsection{Experimental protocols and evaluation metrics}
\label{subsec:exp_pro_eva_met}

Our experiments are designed to simulate a use case of the team-sports video retrieval task (described in Sec.~\ref {sec:introduction}) as mentioned in what follows.

Like query video selection by a user, query videos given at each trial in the experiments must be similar (i.e., must represent the same group activity).
To guarantee that, we use the group activity class annotations provided in team sports datasets for supervised GAR methods.
Although the group activity classes annotated in the datasets are not fine-grained (e.g., spike, set, and pass in volleyball), we utilize them, as with experiments in supervised GAR methods.
Specifically, $N^{\rm query}$ query videos in each retrieval trial are randomly sampled from videos of a single group activity class (denoted by $C^{\rm q}$) in a test set, $\bm{D}^{\rm test}$.
By our query-aware video selection using these $N^{\rm query}$ query videos, $N^{select}$ videos in a training set, $\bm{D}^{\rm train}$, are selected as proposed in Sec.~\ref{subsec:gafl_finetune}.
% 
% Note that these selected videos in $\bm{D}^{\rm train}$ are also annotated with their group activity classes.
These selected videos in $\bm{D}^{\rm train}$ are also annotated with their group activity classes.
Based on the annotated group activity classes in the datasets, we automatically generate binary annotations (i.e., 1 if the annotated group activity class matches $C^{\rm q}$) to simulate user annotations.

After fine-tuning the GAFL network with these query and selected videos with their annotated classes, the retrieval performance is evaluated on each query video as follows.
From each query video, which is originally used as a query video for our query-aware video selection mentioned above, the nearest $K$ videos in the finetuned GAF space are retrieved from a training set, $\bm{D}^{\rm train}$.
%  
%Each retrieved video is automatically classified as positive or negative; if the class of the retrieved video is $C^{\rm q}$, it is labeled as positive.
%   
Then, Precision@K, expressed by $\frac{N^{\rm P}}{K}$ where $N^{\rm P}$ denotes the number of retrieved videos that are correctly classified to $C^{\rm q}$, is evaluated.
Since $N^{\mathrm{select}}$ videos in $\bm{D}^{\rm train}$ are used in the fine-tuning, we use K, which is large enough compared with $N^{\mathrm{select}}$ to evaluate a retrieval performance on videos not used in the fine-tuning.
We also evaluate Hit@K, as used in~\cite{DBLP:conf/eccv/YiSGE22,DBLP:conf/eccv/IbrahimM18}, in which the retrieval is regarded as success (i.e., Hit@K=1) if at least one retrieved video is classified to $C^{\rm q}$.
Otherwise, Hit@K=0.
Hit@K is also suitable to simulate a use case in the team sports retrieval task; team sports analysts may need only one target group activity video among all retrieved videos because the analysts can easily find what they need from the retrieved videos (e.g., three videos) than from all training videos (e.g., thousands of videos).
% 

%For evaluation (Sec.~\ref{subsec:exp_pro_eva_met}), we mainly use Precision@K, measuring the proportion of retrieved videos containing the target group activity. Precision@K is suitable for evaluating this task because collecting videos containing target group activity is important for statistical analysis (e.g., analyzing the success rate of a specific spiking pattern).

In experiments in this paper, the main task is to enhance the performance of retrieval using the original query videos. Evaluation metrics with the original query videos are called ``Precision@K original'' and ``Hit@K original.''
In addition, we also evaluate retrieval using queries sampled from all other test videos (i.e., ``Precision@K others'' and ``Hit@K others'').
This evaluation using the other test videos validates the generalizability of our fine-tuning.

The aforementioned evaluation process is repeated 10 times in each group activity class independently.
That is, in each evaluation process, fine-tuning begins with the original pre-trained model.

\subsection{Datasets}
\label{subsec:datasets}

% The following two datasets of team sports are used to validate the retrieval performance in our experiments.
The following two datasets of team sports (i.e., Volleyball and NBA datasets) are used to validate the retrieval performance in our experiments. To further validate the applicability of our method, we also use the Collective activity dataset as a non-sports dataset.

\noindent{\bf Volleyball dataset}~\cite{DBLP:conf/cvpr/IbrahimMDVM16} consists of 4,830 videos extracted from 55 games. The 4,830 videos are divided into 3,493 train and 1,337 test sets. Each video is annotated with one of the predefined eight group activity classes, i.e., Left-spike, Right-spike, Left-set, Right-set, Left-pass, Right-pass, Left-winpoint, and Right-winpoint.
The detailed information about the group activity annotation data is summarized in Table~\ref{table:det_ann_vol}.
The group activity classes are manually annotated for each video clip. The annotation process is reliable because the annotators can confirm each video, not an image.
While each video has 41 frames, its center frame, nine frames before the center, and ten frames after the center (20 frames, in total) have annotations with the full-body bounding boxes of all players and their action classes, i.e., Waiting, Setting, Digging, Falling, Spiking, Jumping, Moving, Blocking, and Standing. 

\noindent{\bf NBA Dataset (NBA)}~\cite{DBLP:conf/eccv/YanXTS020} comprises 9,172 sequences extracted from 181 professional basketball matches. These sequences are partitioned into a training set of 7,624 sequences and a test set of 1,548 sequences. Each sequence is annotated with one of nine pre-defined group activity classes, including: 2p-succ., 2p-fail.-off., 2p-fail.-def., 2p-layup-succ., 2p-layup-fail.-off., 2p-layup-fail.-def., 3p-succ., 3p-fail.-off., and 3p-fail.-def. 
Table~\ref{table:det_ann_nba} shows the detailed class distribution in the NBA dataset.
The group activity classes are annotated from game logs provided by the NBA's official website. Since abnormal annotations are manually filtered after the initial annotation, the annotation process is reliable, as with the Volleyball dataset.
Unlike the Volleyball dataset, each sequence consists of 72 frames, thereby capturing long-term group activity behaviors.

\noindent{\bf Collective Activity Dataset (CAD)}~\cite{DBLP:conf/iccvw/ChoiSS09} consists of 44 videos.
Every ten frames are annotated with people's bounding boxes and their action classes. Each person's action classes are annotated from either of NA, Crossing, Waiting, Queuing, Walking, or Talking. Different from the Volleyball and NBA datasets, group activity class is determined by the large number of person actions in a frame. 
The detailed distribution of group activity classes is shown in Table~\ref{table:det_ann_cad}.
As the person action classes are manually annotated by confirming the target and the neighboring frames, we can say that the group activity classes obtained from the person action classes are reliable.
For further annotation quality control, we merge Crossing and Walking into Moving due to the high similarity between Crossing and Walking classes, as with ~\cite{DBLP:conf/iccv/Yuan0W21,DBLP:conf/cvpr/WangNY17}.
% 
% We merge Crossing and Walking into Moving as with~\cite{DBLP:conf/iccv/Yuan0W21,DBLP:conf/cvpr/WangNY17}.

\begin{figure*}[t]
  \begin{center}
  \includegraphics[width=\textwidth]{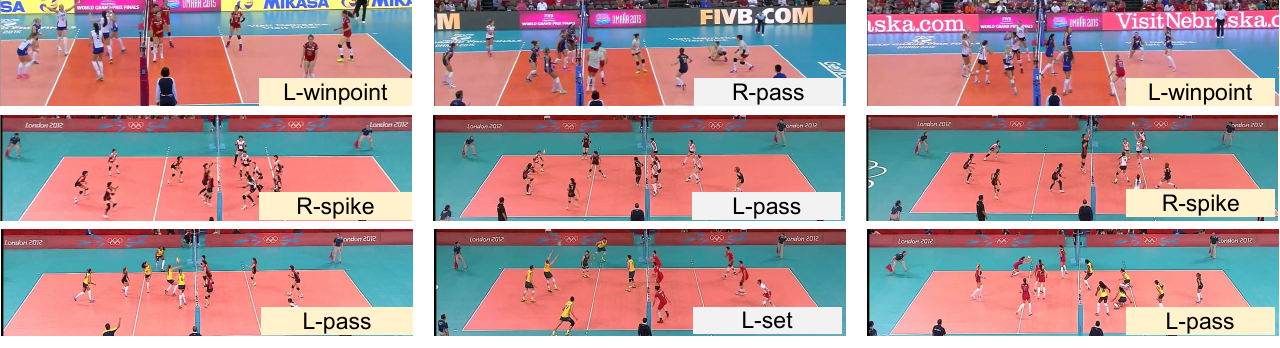}
  Query scene\hspace{45mm}
  GAFL~\cite{DBLP:conf/cvpr/NakataniKU24}\hspace{45mm}
  Ours~\hspace*{5mm}
  \end{center}
  \caption{Comparison of GARet on the Volleyball dataset.
  In all examples, the middle frame of the video is shown for visualization. In each image, manually annotated group activity labels are visualized in the bottom right side of each image.
  }
  \label{fig:comp_ret_vol}
\end{figure*}

\begin{table*}[t]
    \centering
    \caption{
    Detailed analysis of retrieval on the Volleyball Dataset.
    The retrieval results using Precision@10 (original) for each group activity class are shown in the corresponding column. 
    The best result in each column is colored in \red{red}.
    }
    \begin{tabular}{l||c|c|c|c|c|c|c|c} \hline
    Metric = Precision@10 (original) & L-spike & L-pass & L-set & L-winpoint & R-spike & R-pass & R-set & R-winpoint \\ \hline
    B1-Compact128~\cite{DBLP:conf/eccv/IbrahimM18} & 0.320 & 0.243 & 0.357 & 0.330 & 0.307 & 0.340 & 0.260 & 0.223 \\ \hline
    B2-VGG19~\cite{DBLP:conf/eccv/IbrahimM18} & 0.333 & 0.230 & 0.310 & 0.283 & 0.283 & 0.360 & 0.267 & 0.233 \\ \hline
    HRN~\cite{DBLP:conf/eccv/IbrahimM18} & 0.233 & 0.257 & 0.217 & 0.300 & 0.250 & 0.363 & 0.190 & 0.197 \\ \hline
    GAFL~\cite{DBLP:conf/cvpr/NakataniKU24} & 0.763 & 0.393 & 0.557 & 0.473 & 0.777 & 0.467 & 0.450 & 0.303 \\ \hline
    Ours  & \red{0.837} & \red{0.493} & \red{0.683} & \red{0.617} & \red{0.817} & \red{0.613} & \red{0.597} & \red{0.473} \\ \hline
    \end{tabular}
    \label{table:comp_vol_details}
\end{table*}

\subsection{Training Details}

\noindent{\bf Pre-training (step (A) in Fig.~\ref{fig:overview_gafl_fine}):}
Our network is optimized by Adam~\cite{DBLP:journals/corr/KingmaB14} with $\beta_{1}=0.9$, $\beta_{2}=0.999$, and $\epsilon=10^{-8}$. The learning rate is 0.0001. 
% In both the Volleyball and NBA datasets, the whole image is resized to 320x640. 
In all datasets, the whole image is resized to 320x640. 
We employ the VGG-19 models as a person feature extractor (Fig.~\ref{fig:overview_network} (a)).
Our $\mathrm{AFH}$ consists of three fully-connected layers. 

\noindent{\bf Fine-tuning (step (B) in Fig.~\ref{fig:overview_gafl_fine}):}
In our human-in-the-loop adaptation, our network is also optimized by Adam $\beta_{1}=0.9$, $\beta_{2}=0.999$, and $\epsilon=10^{-8}$ with the same parameters as the pre-training stage. 
The margin $\alpha$ in contrastive learning is set to 10.
$N^{\rm query} = 3$ and $N^{\rm select} = 5$ in our experiments.
% 
% In both the Volleyball and NBA datasets, 
In all datasets, 
$N^{E} = 4$ and $N^{V} = 2$ are used in our proposed query-aware video selection (Sec.~\ref{subsubsec:select_query}).

% \begin{table}[t]
%     \centering
%     \caption{
%     Quantitative comparison of retrieval on the VolleyBall Dataset (Volleyball).
%     The best result in each column is colored in \textcolor{red}{red}.
%     In ``Ours'', the whole network is fine-tuned for each target group activity with 5 videos for fine-tuning. As query videos, 3 videos are sampled from the test videos for each fine-tuning. 
%     The retrieval results are the weighted average of 8 group activity classes, considering the number of videos in each group activity class.}
%     \begin{tabular}{l||c|c} \hline
%     & Precision@5 & Precision@10 \\ \hline
%     B1-Compact128~\cite{DBLP:conf/eccv/IbrahimM18} & 0.318 & 0.299 \\ \hline
%     B2-VGG19~\cite{DBLP:conf/eccv/IbrahimM18} & 0.319 & 0.291 \\ \hline
%     HRN~\cite{DBLP:conf/eccv/IbrahimM18} & 0.280 & 0.254 \\ \hline
%     GAFL~\cite{DBLP:conf/cvpr/NakataniKU24} & 0.557 & 0.533 \\ \hline
%     Ours  & \red{0.739} & \red{0.647} \\ \hline
%     \end{tabular}
%     \label{table:comp_vol}
% \end{table}

\if 0
\begin{figure*}
    \centering
    \includegraphics[width=0.95\textwidth]{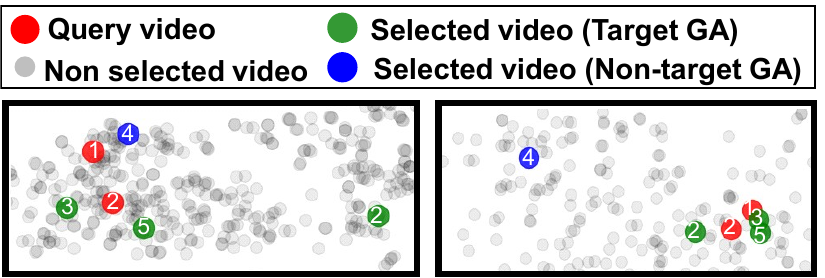}
    \hspace*{-10mm}
    GAFL~\cite{DBLP:conf/cvpr/NakataniKU24}~\hspace{80mm}
    Ours
    \caption{
    Visualization of GAFs extracted from the query and selected videos in our method.
    The target group activity is ``L-set'' in this figure.
    $2C$-dimensional GAF obtained from each video is embedded into a 2D space by t-SNE. 
    }
    \label{fig:comp_vol_gaf_tsne_det}
\end{figure*}
\fi

\begin{figure}
    \centering
    \includegraphics[width=0.95\columnwidth]{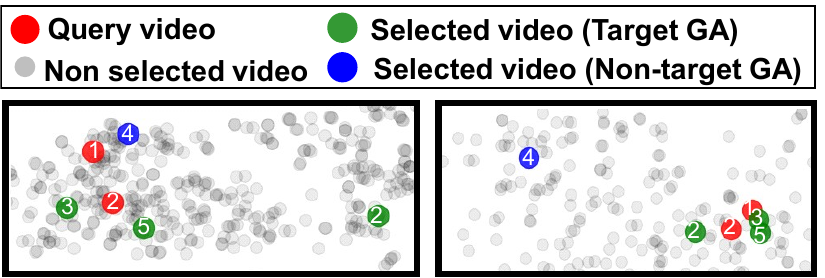}
    \hspace*{-5mm}
    GAFL~\cite{DBLP:conf/cvpr/NakataniKU24}~\hspace{30mm}
    Ours
    \caption{
    % Visualization of GAFs extracted from the query and selected videos in our method.
    % $2C$-dimensional GAF obtained from each video is embedded into a 2D space by t-SNE. 
    Visualization of GAFs extracted from the query and selected videos in our method by t-SNE.
    The target group activity is ``L-set'' in this figure.
    In this GAF visualization, indices of the query and selected videos are shown in each corresponding GAF.
    % In this GAF visualization, indices of the query videos (i.e., 1, 2, and 3) and indices of the selected videos (i.e., 1, 2, 3, 4, and 5) are shown with each corresponding GAF.
    % 
    While we utilize three query videos and five selected videos in our experiments, we display only two query videos and four selected videos due to space constraints. 
    % 
    % In the upper column, the middle frame of the 2nd, 4th, and 5th selected videos is shown, along with their index on the bottom right side.
    }
    \label{fig:comp_vol_gaf_tsne_det}
\end{figure}

\begin{figure}
    \centering
    \includegraphics[width=0.95\columnwidth]{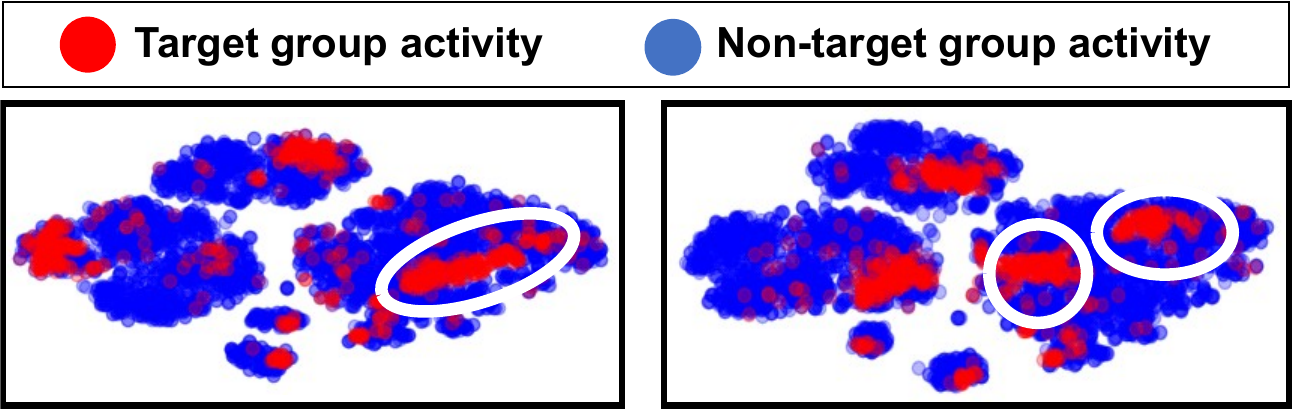}
    \hspace*{-5mm}
    GAFL~\cite{DBLP:conf/cvpr/NakataniKU24}~\hspace{30mm}
    Ours
    \caption{Visualization of learned GAF space by t-SNE on the Volleyball dataset.
    The target group activity is ``L-set'' in this figure.
    % $2C$-dimensional GAF obtained from video is embedded into a 2D space for visualization. 
    % The color of each data point shows a target group activity label corresponding to each test video.
    }
    \label{fig:comp_vol_gaf_tsne}
\end{figure}

\begin{figure*}[t]
  \begin{center}
  \includegraphics[width=\textwidth]{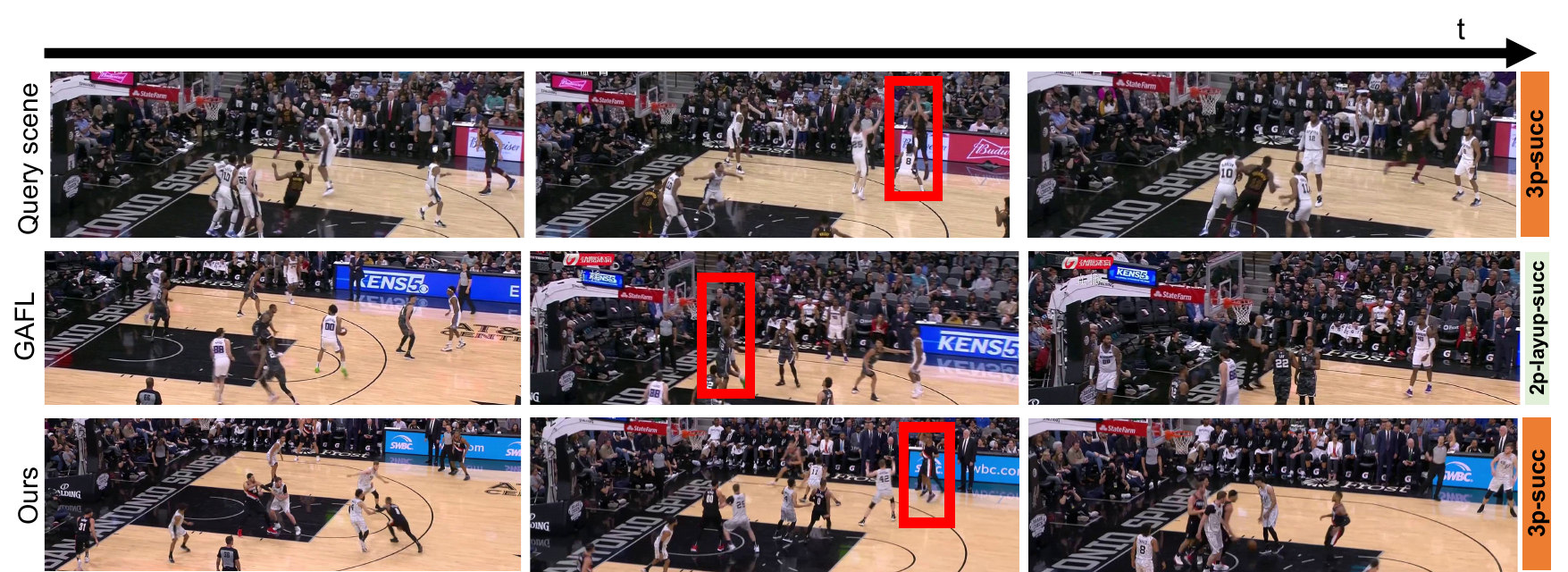}
  \end{center}
  \caption{Comparison of GARet on the NBA dataset.
  % The group activity label is shown on the rightmost side of each scene.
  % Multiple images from a video are shown in each column.
  % 
  A shooter is enclosed by a red rectangle in the images.
  In each example, manually annotated group activity labels are visualized on the rightmost side.
  }
  \label{fig:comp_ret_nba}
\end{figure*}

\subsection{Comparative Experiments}
\label{subsec:comp_exp}

Based on the experimental protocols and evaluation metrics (Sec.~\ref{subsec:exp_pro_eva_met}), we compare our method with several methods on the Volleyball and NBA datasets as described in Sec.~\ref{subsec:datasets}. In this section, we comprehensively compare these methods through qualitative and quantitative experiments for each dataset.

Our method is compared with two baseline methods shown in~\cite{DBLP:conf/eccv/IbrahimM18} and two SOTA methods~\cite{DBLP:conf/eccv/IbrahimM18,DBLP:conf/cvpr/NakataniKU24}.
In the two baseline methods and~\cite{DBLP:conf/eccv/IbrahimM18}, the set of person features (i.e., the concatenation of $\bm{F}^{\mathrm{mask}}_{\mathrm{TS}}$ and $\bm{F}^{\mathrm{mask}}_{\mathrm{ST}}$ in Fig.~\ref{fig:overview_network} (b)) is used for retrieval, while $\bm{G}$ is used in GAFL~\cite{DBLP:conf/cvpr/NakataniKU24} and our method.
The essential difference between our method and all the others is that our method fine-tunes the GAF model in a human-in-the-loop manner.

\noindent{\bf Volleyball dataset.}
Table~\ref{table:comp_vol} shows that our method significantly improves the retrieval performance.
We can see that the retrieval results are improved not only for the original query videos used for fine-tuning but also for the other videos.
% 
%These results validate that our proposed fine-tuning can improve the retrieval performance in a generalized manner.

Figure~\ref{fig:comp_ret_vol} shows visual retrieval results.
In each row of the left column, a middle frame of each query video is shown.
In the center and right columns, the middle frames of videos retrieved by GAFL and our method (i.e., fine-tuned GAFL) are shown, respectively.
% 
%%%Each retrieval is conducted with the nearest neighbor search.

In the upper example, our method successfully retrieves ``L-winpoint,'' while GAFL~\cite{DBLP:conf/cvpr/NakataniKU24} retrieves ``R-pass.''
This difference can be interpreted as our method learns the characteristic configuration of people on the left side (i.e., people are celebrating a point in a circle).
The middle example also shows that our method emphasizes people related to the spike group activity (e.g., spiking and blocking person actions) in the GAF during fine-tuning.
In the bottom example, GAFL~\cite{DBLP:conf/cvpr/NakataniKU24} incorrectly retrieves ``L-set'' from ``L-pass'' query videos. 
This failure result may come from the similar configurations of people between the query and the retrieved videos. 
That is, there are three people who are standing in front of the net on the right side in both the query video and the video retrieved by GAFL.
Different from GAFL, our method correctly retrieves ``L-pass.''
This result indicates that our proposed video selection process selects various L-pass videos with different people configurations for fine-tuning.

Retrieval results on each group activity are shown in Table~\ref{table:comp_vol_details}. 
We can see that our method works well in all group activities. 
These results validate that our proposed fine-tuning can be generalized for any group activity.

\begin{figure*}[t]
\begin{center}
\includegraphics[width=\textwidth]{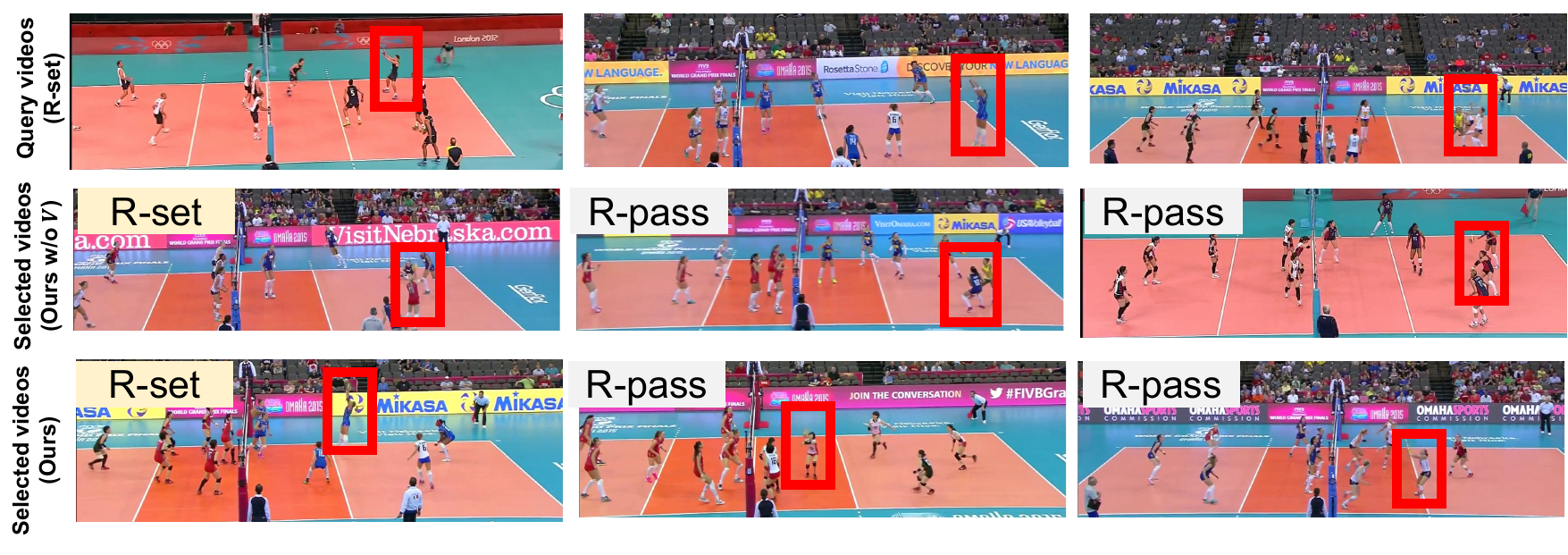}
\end{center}
\caption{
Examples of videos selected in ``Ours w/o $\bm{V}$'' and ``Ours'' when the queries are R-set videos. 
The upper row shows the query videos. 
In the second and third rows, the videos selected by ``Ours w/o $\bm{V}$'' and ``Ours'' are shown with their manually annotated group activity labels, respectively.
A receiving player is enclosed by a red rectangle in the images.
Due to space limitations, only three videos are displayed from the selected $N^{\mathrm{select}}$ videos.
}
\label{fig:exa_sec}
\end{figure*}

\begin{table}[t]
    \centering
    \caption{
    Quantitative comparison of retrieval on the NBA Dataset.
    The best result in each column is colored in \textcolor{red}{red}.
    In ``Ours'', the whole network is fine-tuned for each target group activity with five selected videos. As query videos, three videos are sampled from the test videos for each fine-tuning. 
    % The retrieval results are the weighted average of 8 group activity classes, considering the number of videos in each group activity class.
    }
    \begin{tabular}{l||c|c} \hline
    % & Precision@5 & Precision@10 \\ 
    % & original & original \\ \hline
    % B1-Compact128~\cite{DBLP:conf/eccv/IbrahimM18} & 0.159 & 0.161 \\ \hline
    % B2-VGG19~\cite{DBLP:conf/eccv/IbrahimM18} & 0.160 & 0.156 \\ \hline
    % HRN~\cite{DBLP:conf/eccv/IbrahimM18} & 0.137 & 0.135 \\ \hline
    % GAFL~\cite{DBLP:conf/cvpr/NakataniKU24} & 0.215 & 0.206 \\ \hline
    % Ours  & \red{0.258} & \red{0.233} \\ \hline
    & Precision@10 & Hit@10 \\ 
    & original & original \\ \hline
    B1-Compact128~\cite{DBLP:conf/eccv/IbrahimM18} & 0.161 & 0.749 \\ \hline
    B2-VGG19~\cite{DBLP:conf/eccv/IbrahimM18} & 0.156 & 0.776 \\ \hline
    HRN~\cite{DBLP:conf/eccv/IbrahimM18} & 0.135 & 0.771 \\ \hline
    GAFL~\cite{DBLP:conf/cvpr/NakataniKU24} & 0.206 & 0.798 \\ \hline
    Ours & \red{0.233} & \red{0.839} \\ \hline
    \end{tabular}
    \label{table:comp_nba}
\end{table}

Figure~\ref{fig:comp_vol_gaf_tsne_det} shows the GAFs extracted from the query and selected videos visualized by t-SNE~\cite{van2008visualizing} on the Volleyball dataset. 
The GAF space is fine-tuned for ``L-set.''
% The GAFs extracted from the query and selected videos are visualized with their middle image.
The GAFs extracted from the query and selected videos are visualized with the middle frame of each video. 
% 
% Compared to GAFL~\cite{DBLP:conf/cvpr/NakataniKU24}, ``Ours,'' successfully projects the $2$nd, $3$rd, and $4$th selected videos annotated as target group activity (indicated by green points) closer to the $1$st and $2$nd query videos.
Compared to GAFL~\cite{DBLP:conf/cvpr/NakataniKU24}, ``Ours'' successfully projects three selected videos annotated as target group activity (indicated by green points) closer to the query videos.
%  
% However, even in 'Ours,' the $3$rd query video is far from the other query and selected videos, except for the $1$st selected video.
% 
% The results can be interpreted as follows: since the location of the setting player in the $3$rd query video differs from that in the other query and selected videos, it is not easy to update the GAF to be close to these videos as L-set activity. Regarding the $1$st selected video, the location of the setting player is similar to the $3$rd query video, and their GAFs are projected into the near region.

Figure~\ref{fig:comp_vol_gaf_tsne} shows the learned GAF spaces on the Volleyball dataset. 
Different from Fig.~\ref{fig:comp_vol_gaf_tsne_det}, GAFs extracted from all videos in $\bm{D}^{test}$ are visualized in this figure.
We can see that the GAFs of the target group activity videos (indicated by red points) are distributed widely in GAFL~\cite{DBLP:conf/cvpr/NakataniKU24}, as shown in the white circle.
Compared to this GAF space of GAFL, in the GAF space fine-tuned by our method, the GAFs of the target group activity videos are projected within smaller regions as enclosed by the two white circles.
% 
% The results indicate that our fine-tuning generally updates the GAF space to improve discriminativity for the target group activity beyond the limited GAF feature regions around query videos, as also proved in Table~\ref{table:comp_vol}.
The results indicate that our fine-tuning generally updates the GAF space to improve discriminativity for the target group activity in all videos in  $\bm{D}^{test}$, as also proved by “Precision@K others” and “Hit@K others” in Table~\ref{table:comp_vol}.

\noindent{\bf NBA dataset.}
The comparison of GARet on the NBA dataset is shown in Table~\ref{table:comp_nba}. 
Table~\ref{table:comp_nba} shows that our method is the best in all metrics on the NBA dataset. These results demonstrate the generalization ability of our method across various team sports, which differ in terms of the number of people and the granularity of target group activities.

Figure~\ref{fig:comp_ret_nba} shows visual retrieval results on the NBA dataset. In the upper row, the query videos used for retrieval are shown with their group activity labels. In the second and third rows, videos retrieved by GAFL~\cite{DBLP:conf/cvpr/NakataniKU24} and our method are shown with their group activity labels, respectively.
A shooter is enclosed by a red rectangle in the images.
In this example, the video annotated as 3p-succ activity is used as the query. While the video in which 2p-layup-succ activity is observed is retrieved by GAFL~\cite{DBLP:conf/cvpr/NakataniKU24}, our method successfully retrieves the video in which 3p-succ activity is observed.

\begin{table}[t]
    \centering
    \caption{
    Quantitative comparison of retrieval on the Collective Activity Dataset.
    The best and second-best results in each column are colored in \textcolor{red}{red} and \textcolor{blue}{blue}.
    In ``Ours'', the whole network is fine-tuned for each target group activity with five selected videos. As query videos, three videos are sampled from the test videos for each fine-tuning. 
    }
    \begin{tabular}{l||c|c} \hline
    & Precision@10 & Hit@10 \\ 
    & original & original \\ \hline
    B1-Compact128~\cite{DBLP:conf/eccv/IbrahimM18} & \blue{0.799} & \red{0.925} \\ \hline
    B2-VGG19~\cite{DBLP:conf/eccv/IbrahimM18} & 0.660 & 0.901 \\ \hline
    HRN~\cite{DBLP:conf/eccv/IbrahimM18} & 0.522 & 0.859 \\ \hline
    GAFL~\cite{DBLP:conf/cvpr/NakataniKU24} & 0.788 & 0.840 \\ \hline
    Ours & \red{0.814} & \blue{0.866} \\ \hline
    \end{tabular}
    \label{table:comp_cad}
\end{table}

\noindent{\bf Collective Activity Dataset.}
In addition to validating the two sports datasets, which is the main focus of this work, we also conduct experiments on the Collective Activity Dataset. This dataset includes general living scenes and can be used to validate the generality of our method.

Table~\ref{table:comp_cad} shows the comparison of GARet on the Collective Activity Dataset. In the experiments on this dataset, we empirically decided to remove $\mathcal{L}_{reg}$ from $\mathcal{L}$.
We can see that our method is best in ``Precision@10 original'' in Table~\ref{table:comp_cad}. These results demonstrate that GARet by our method is applicable to non-sports videos. While ``B1-Compact128'' is better than ``Ours'' in Hit@10 original, the result in our method is certainly improved from the result in the pre-trained model (i.e., GAFL~\cite{DBLP:conf/cvpr/NakataniKU24}). The performance gap can be filled with improvements in the pre-trained models.

\subsection{Detailed analysis}
\label{subsec:det_analysis}

\begin{table}[t]
    \centering
    \caption{
    Comparison of video selection on the Volleyball dataset.
    The best result in each column is colored in \textcolor{red}{red}.
    }
    \begin{tabular}{l||c|c} \hline
    & Precision@10 & Hit@10 \\
    & original & original \\ \hline
    Random & 0.565 & 0.546 \\ \hline
    K-means~\cite{DBLP:journals/corr/abs-1901-05954}& 0.568 & 0.550 \\ \hline
    Core-set~\cite{DBLP:conf/iclr/SenerS18} & 0.560 & 0.543 \\ \hline
    Ours w/o $\bm{S}$ & 0.566 & 0.545 \\ \hline
    Ours w/o $\bm{V}$ & 0.640 & 0.562 \\ \hline
    Ours & \red{0.647} & \red{0.564} \\ \hline
     \end{tabular}
    \label{table:comp_vid_sec_vol}
\end{table}

\subsubsection{Comparison of video selection}
\label{subsubsection:com_vid_select}

To validate the effectiveness of our proposed video selection, we compare the retrieval results with GAF models fine-tuned by videos selected using the following five other methods, as shown in Table~\ref{table:comp_vid_sec_vol}. 
In ``Random,'' videos are randomly selected from $\bm{D^{\mathrm{train}}}$. In ``Core-set'' and ``K-means,'' GAFs extracted by the pre-trained GAFL network are used for each method.
In ``Ours w/o $\bm{S}$'' and ``Ours w/o $\bm{V}$,'' $\bm{S}$ and $\bm{V}$ are removed from $\bm{I}$ (shown in Eq.~\ref{eq:q_info}), respectively.

The results validate that our method is best in all metrics. 
Compared with ``Core-set'' and ``K-means'' to our method, our method is significantly better (i.e., while Precision@10 (original) is 0.560 and 0.568 in ``Core-set'' and ``K-means, it is 0.647 in our method). 
% 
% These results indicate that our proposed query-aware video selection successfully selects videos which is effective for this fine-tuning.
These results indicate that our proposed query-aware video selection for selecting valuable videos is effective compared to other data selections in this fine-tuning.
% 
% Furthermore, the performance gain of ``Ours'' from ``Ours w/o $\bm{V}$'' demonstrated that query local dissimilarity contributes to video selection, which is important to learn discriminative features.
Furthermore, the performance gain of ``Ours'' from ``Ours w/o $\bm{V}$'' demonstrated that query local dissimilarity contributes to video selection. The results demonstrated that the motivation of $\bm{V}$ (i.e., selecting videos locally dissimilar to the query videos) is important to learn discriminative features.
We can also see the performance gain from ``Ours w/o $\bm{S}$.'' 
This result validates that query similarity (i.e., $\bm{S}$, as defined in Eq.~\ref{eq:q_sim}) in our method improves the retrieval performance of target group activity.
Since $\bm{S}$ is proposed for selecting videos globally similar to the query videos, the results indicate that selecting globally similar videos to the query videos is important for this fine-tuning in addition to selecting the local dissimilar videos by $\bm{V}$, as defined in Eq.~\ref{eq:q_var}.

\subsubsection{Example of selected videos}
\label{subsubsec:exa_sec}

Figure~\ref{fig:exa_sec} shows the example of videos selected in ``Ours'' and ``Ours w/o $\bm{V}$'' when the queries are R-set videos on the Volleyball dataset.
While both ``Ours'' and ``Ours w/o $\bm{V}$'' select videos in which a ball is received by a player (enclosed by a red rectangle), their locations are different.
In ``Ours w/o $\bm{V}$,'' the location of the player receiving the ball in each video is similar (i.e., the player is located on the rear side of the right court). Unlike ``Ours w/o $\bm{V}$,'' ``Ours'' selects videos where the receiving player is located in various areas on the court.
Using the selected videos in which players receive the ball in different locations, ``Ours'' efficiently learns the GAF discriminative for the R-set group activity.

\if 0
Figure~\ref{fig:exa_sec_rwin} shows examples of selected videos when the queries are R-set videos on the Volleyball dataset. While the same two videos are selected in ``Ours w/o $\bm{V}$'' and ``Ours,'' we can see that ``Ours'' selects the R-pass video in which the people configuration on the right side of the court (enclosed by a red rectangle) is similar to that of R-winpoint videos. The R-pass video that is subtly similar to R-winpoint videos is useful to fine-tune the GAF space for a target group activity (i.e., R-winpoint).
\fi

\begin{figure}
    \centering
    \includegraphics[width=\columnwidth]{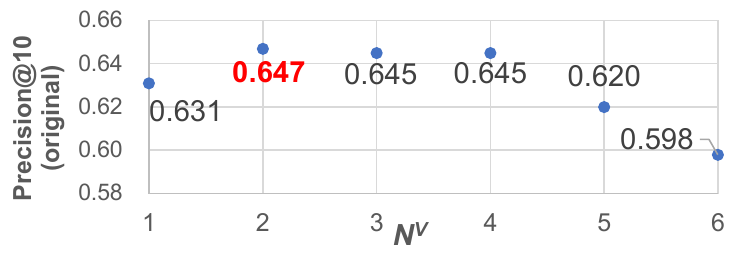}
    \caption{
    Retrieval performance changes depending on $N^{V}$ in query local dissimilarity on the Volleyball dataset.
    $N^{V}$ is increased from 1 to 6 in increments of 1.
    Retrieval results evaluated by Precision@10 (original) are shown in this figure. 
    }
    \label{fig:comp_nv_vol}
\end{figure}

\subsubsection{Analysis of query-aware video selection}

\noindent{\bf Effect of $N^{V}$.}
% Figure~\ref{fig:comp_nv_vol} shows the retrieval performance when $N^{V}$ (i.e., the number of masking people) is changed in perturbed query-similarity variance (Sec.~\ref{subsubsec:select_query}) on the Volleyball dataset. 
Figure~\ref{fig:comp_nv_vol} shows the retrieval performance changes depending on $N^{V}$ (i.e., the number of masking people) in query local dissimilarity (Sec.~\ref{subsubsec:select_query}) on the Volleyball dataset.
In this figure, $N^{V}$ is increased from 1 to 6 in increments of 1.
We can see that the best result is obtained with $N^{V}=2$, and the results significantly drop with $N^{V}> 5$.
These results validate that masking using a small number of people is effective in our query local dissimilarity.

\begin{figure}
    \centering
    \includegraphics[width=\columnwidth]{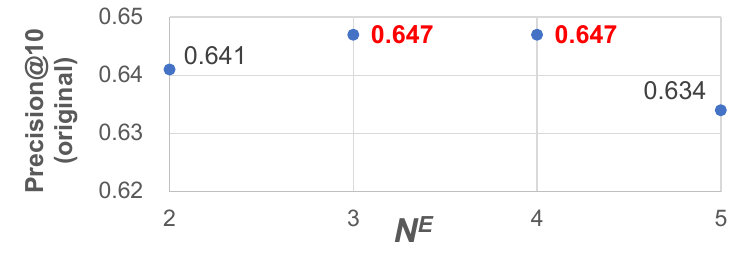}
    \caption{
    Retrieval performance changes depending on  $N^{E}$ in query-aware video selection on the Volleyball dataset.
    $N^{E}$ is increased from 2 to 5 in increments of 1.
    Retrieval results evaluated by Precision@10 (original) are shown in this figure. 
    }
    \label{fig:comp_ne_vol}
\end{figure}

\noindent{\bf Effect of $N^{E}$.}
As mentioned in Sec.~\ref{subsubsec:select_query}, $N^{E}$ controls the number of videos selected in our query-aware video selection before diversity-aware video selection described in Sec.~\ref{subsubsec:select_div}.
Figure~\ref{fig:comp_ne_vol} shows that the retrieval performance changes on the Volleyball dataset when $N^{E}$ is changed from 2 to 5 in increments of 1.
We can see that the best results are obtained when $N^{E}=3,4$.
Furthermore, we can see that the retrieval performance is largely dropped with $N^{E}=5$ (i.e., 0.647 in $N^{E}=4$ drops to 0.634 in $N^{E}=5$).
The results validate that selecting videos far from the query videos in the pre-trained GAF space degrades the retrieval performance.

%%%%%%%%%%%%%%%%%%%%%%%%%%%%%%%%%%%%%%%%%%%%%%%%%%%%%%%%%%%%%%%%%%%%%%

\section{Concluding Remarks}
\label{sec:conclusion}
%%% Summary
This paper proposed a human-in-the-loop adaptation in the pre-trained GAFL for team sports video retrieval.
The pre-trained GAFL is updated for the target group activity presented by query videos given by users.
In addition, query-aware video selection specialized for GARet is also proposed for data efficiency to reduce users' annotations.
While weak supervision (i.e., positive and negative labeling to selected videos) is additionally required compared with the fully-unsupervised baseline~\cite{DBLP:conf/cvpr/NakataniKU24}, our proposed human-in-the-loop adaptation significantly improves the retrieval performance on two team sports datasets (e.g., the performance gain is 0.114 in Precision@10 original compared ``Ours'' with GAFL~\cite{DBLP:conf/cvpr/NakataniKU24} on the Volleyball dataset).

%%% Future work
% While our method is better than other GAF learning methods, it requires weak annotations in a human-in-the-loop manner. 
% 
Future research directions to further decrease annotation costs include explicitly identifying key people of the target group activity, which is implicitly learned from a large number of videos.

%% The Appendices part is started with the command \appendix;
%% appendix sections are then done as normal sections
% \appendix
% \section{Example Appendix Section}
% \label{app1}

% Appendix text.

%% For citations use: 
%%       \cite{<label>} ==> [1]

%%
% Example citation, See \cite{lamport94}.

%% If you have bib database file and want bibtex to generate the
%% bibitems, please use
%%
 \bibliographystyle{elsarticle-num} 
 \bibliography{main}

%% else use the following coding to input the bibitems directly in the
%% TeX file.

%% Refer following link for more details about bibliography and citations.
%% https://en.wikibooks.org/wiki/LaTeX/Bibliography_Management

% \begin{thebibliography}{00}

%% For numbered reference style
%% \bibitem{label}
%% Text of bibliographic item

% \bibitem{lamport94}
%   Leslie Lamport,
%   \textit{\LaTeX: a document preparation system},
%   Addison Wesley, Massachusetts,
%   2nd edition,
%   1994.

% \end{thebibliography}
\end{document}